\documentclass[final,1p,times]{elsarticle}

\usepackage{graphicx}
\usepackage{float}
\usepackage{amsmath}
\usepackage{booktabs} 
\usepackage{algorithm}
\usepackage{algpseudocode}
\usepackage{hyperref}
\usepackage{float}

\usepackage{amssymb}
\journal{}

\begin{document}

\begin{frontmatter}

\title{Bayesian Optimized Deep Ensemble for Uncertainty Quantification of Deep Neural Networks: a System Safety Case Study on Sodium Fast Reactor Thermal Stratification Modeling}

\author[inst1]{Zaid Abulawi}
\ead{zaidabulawi@tamu.edu}

\author[inst2]{Rui Hu}
\ead{rhu@anl.gov}

\author[inst3]{Prasanna Balaprakash}
\ead{pbalapra@ornl.gov}

\author[inst1]{Yang Liu}
\ead{y-liu@tamu.edu}

\affiliation[inst1]{organization={Texas A\&M University, Department of Nuclear Engineering},
            country={USA}}

\affiliation[inst2]{organization={Argonne National Laboratory, Nuclear Science and Engineering Division},
            country={USA}}

\affiliation[inst3]{organization={Oak Ridge National Laboratory, Computing and Computational Sciences Directorate},
            country={USA}}

\begin{abstract}
Accurate predictions and uncertainty quantification (UQ) are essential for decision-making in risk-sensitive fields such as system safety modeling. Deep ensembles (DEs) are efficient and scalable methods for UQ in Deep Neural Networks (DNNs); however, their performance is limited when constructed by simply retraining the same DNN multiple times with randomly sampled initializations. To overcome this limitation, we propose a novel method that combines Bayesian optimization (BO) with DE, referred to as BODE, to enhance both predictive accuracy and UQ.

We apply BODE to a case study involving a Densely connected Convolutional Neural Network (DCNN) trained on computational fluid dynamics (CFD) data to predict eddy viscosity in sodium fast reactor thermal stratification modeling. Compared to a manually tuned baseline ensemble, BODE estimates total uncertainty approximately four times lower in a noise-free environment, primarily due to the baseline's overestimation of aleatoric uncertainty. Specifically, BODE estimates aleatoric uncertainty close to zero, while aleatoric uncertainty dominates the total uncertainty in the baseline ensemble. We also observe a reduction of more than 30\% in epistemic uncertainty. When Gaussian noise with standard deviations of 5\% and 10\% is introduced into the data, BODE accurately fits the data and estimates uncertainty that aligns with the data noise. These results demonstrate that BODE effectively reduces uncertainty and enhances predictions in data-driven models, making it a flexible approach for various applications requiring accurate predictions and robust UQ.
\end{abstract}

\begin{keyword}
Bayesian Optimization, Uncertainty Quantification, Deep Ensemble, Data-Driven Turbulence Modeling, Data Noise
\end{keyword}

\end{frontmatter}

\section{Introduction}

\subsection{Literature Review}

Accurate predictions with reliable uncertainty quantification (UQ) are essential in risk-sensitive fields such as aerodynamics, medicine, material science, and nuclear safety. Effective UQ allows decision-makers to assess the confidence in model predictions, ensuring reliability when applied to critical systems and components. In recent years, Deep Neural Networks (DNNs) have been increasingly employed across various domains to enhance prediction capabilities and provide UQ.

In material science, for instance, Bansal et al.~\cite{bansal2022physics} utilized a Physics-Informed Neural Network combined with Gaussian Processes (GP) to predict and quantify uncertainties in material loss due to corrosion in aluminum-iron joints. Similarly, Rathnakumar et al.~\cite{rathnakumar2023epistemic} developed a Bayesian Boundary-Aware Convolutional Network for crack detection in materials, offering both predictive capability and uncertainty estimates. In the field of energy storage technology, Lin et al.~\cite{lin2023battery} assessed uncertainties in predicting the state of health of lithium-ion batteries using a Gated Recurrent Unit Neural Network paired with a Hidden Markov Model.

Engineering and infrastructure management have also benefited from DNNs. Fan et al.~\cite{FAN2023109088} used Long Short-Term Memory (LSTM) models combined with Monte Carlo methods to predict failure rates in water distribution networks, capturing uncertainties in network deterioration. In biomechanics, Kakhaia et al.~\cite{kakhaia2023inverse} applied inverse UQ and GP to approximate an agent-based model simulating arterial tissue mechanics, providing a robust framework for understanding tissue behavior.

In nuclear engineering, DNNs and other machine learning (ML) methods have demonstrated great potential in a variety of applications, ranging from reactor optimization and control~\cite{radaideh2021neorl, SUN2024112312}, transient prediction~\cite{guillen2020relap5}, model validation~\cite{liu2019validation, xie2021towards, liu2019validation-2}, risk analysis~\cite{junyung2021system}, to the development of digital twins~\cite{lin2021dt, liu2024development, gong2023parameter, kobayashi2024explainable}. Particularly, DNNs have shown excellent performance in learning highly non-linear relationships and bridging gaps between multiscale datasets~\cite{dave2024physics, liu2018data, bao2020using}, making them a promising tool for parametric partial differential equations (PDEs). Techniques like Principal Component Analysis (PCA), DNNs, and GP have been integrated to improve the reliability of multiphase Computational Fluid Dynamics (CFD) bubbly flow simulations~\cite{liu2021uncertainty_b}. Moreover, for critical heat flux prediction, conditional variational autoencoders, DNNs, and Deep Ensembles (DEs) have been utilized to ensure accurate predictions while addressing data uncertainties~\cite{alsafadi2024predicting, zhao2020prediction}. Applying ML to turbulence modeling is also an active research area with many recent advances~\cite{ling2016reynolds, duraisamy2019turbulence} that benefit nuclear thermal hydraulics modeling. Specifically, we developed a successful application that leverages DNNs to create a coarse-mesh eddy viscosity closure supported by fine-mesh CFD data~\cite{liu2022data}, which serves as the baseline model for this work.

For most DNN-related tasks in engineering applications, we often require not just accurate model predictions but also the uncertainties associated with those predictions to make better decisions based on the DNN outputs. UQ for classical nuclear reactor system modeling and simulation is an active research topic, with multiple methods developed and successfully applied—particularly those focusing on using Bayesian inference for inverse UQ to reduce and quantify model discrepancies with experimental measurements~\cite{liu2019uncertainty, kobayashi2024ai, wu2018inverse_1, radaideh2019integrated}. However, applying UQ to DNN models for specific engineering problems remains a less-studied area.

Various UQ methods for DNN models have been developed, each with its strengths and limitations depending on the application. Bayesian Neural Networks (BNNs) provide uncertainties through Bayesian inference but are computationally expensive and challenging to scale~\cite{nemani2023uncertainty, dabiran2023sparse}. Monte Carlo (MC) dropout is a faster alternative, applying dropout layers at prediction time, but it may not perform well on complex neural networks~\cite{nemani2023uncertainty}. DEs offer scalability, high predictive accuracy, and high-quality UQ capabilities~\cite{nemani2023uncertainty}. They have been suggested to improve generalization, reduce overfitting, and quantify both aleatoric (uncertainty due to the stochastic nature of the system) and epistemic (uncertainty due to lack of knowledge) uncertainties~\cite{yaseen2023quantification, lakshminarayanan2017simple}. However, DEs trained with fixed DNN setup with only different weight/bias initializations demonstrated less satisfactory results for complicated problems~\cite{egele2022autodeuq}.

The performance of a DNN is significantly influenced by its hyperparameters, such as the number of layers, activation functions, and learning rates~\cite{raiaan2024systematic}. Hyperparameter optimization involves identifying the set of hyperparameters that result in a model with high accuracy in a reasonable time~\cite{aszemi2019hyperparameter}. Model-free optimization algorithms like grid search, random search, and population-based methods (genetic algorithms, evolutionary algorithms, particle swarm optimization) do not rely on surrogate models~\cite{tuba2021convolutional}. In contrast, model-based hyperparameter optimization, such as Bayesian Optimization (BO), uses a surrogate model to guide the search for optimal hyperparameters~\cite{tuba2021convolutional, dabiran2023sparse}.

Optimizing DNNs is crucial to minimize discrepancies between targets and predictions and to improve UQ analyses. Recent two pioneering studies have demonstrated that integrating optimization techniques with DNNs can significantly enhance both prediction accuracy and UQ for complicated problems~\cite{egele2022autodeuq, balaprakash2018deephyper}. Previous research on DEs for UQ has explored various strategies for model selection and optimization. For example, Wenzel et al.~\cite{wenzel2020hyperparameter} used random search to generate a diverse pool of models, followed by a greedy selection algorithm to identify the best subset for an ensemble. Another study by Lu et al.~\cite{lu2024deep} enhanced UQ with DEs of Convolutional Neural Networks (CNNs) by optimizing the number of neurons and kernels using the Hyperband algorithm and fine-tuning the learning rate with a warm-up algorithm.

BO has proven effective for various applications and neural network architectures, including health~\cite{health}, economics~\cite{econommics}, environmental science~\cite{environmental}, material science~\cite{UENO201618}, and physics and topology optimization~\cite{Kim2021DeepLF}. The computational cost of BO can be reduced by limiting the number of hyperparameters being optimized, narrowing the range of hyperparameter values, reducing the number of evaluations, applying constrained BO, or using mathematical techniques to cover the search space when defining the prior distribution in GP.

\subsection{Data-Driven Eddy Viscosity in SAM}

One of the key domains where accurate prediction and high-quality UQ are crucial is nuclear system safety. Advanced nuclear reactors differ in design from traditional nuclear power plants, aiming for enhanced safety and economics. Modern and adaptable modeling tools are needed to ensure the safety of these new designs and their associated components. The System Analysis Module (SAM) is a system-level code developed at Argonne National Laboratory~\cite{hu2021sam}, designed for the analysis of advanced reactor systems.

Although system-level codes primarily conduct one-dimensional (1-D) simulations, three-dimensional (3-D) simulations are necessary for certain components to analyze complex thermal-fluid behaviors. For example, 3-D analysis is essential for thermal stratification and fluid mixing phenomena in large pools or tanks, as occurs in pool-type sodium-cooled fast reactors~\cite{lu2020sensitivity, liu2023benchmarking}, suppression pools of Boiling Water Reactors~\cite{bao2018safe}, and components relying on natural circulation for cooling~\cite{liu2022data}. Current system codes lack accurate stratification and thermal mixing models, which are critical for safety assessments.

SAM incorporates a 3-D module to model these thermal-fluid phenomena using a coarse mesh setup to maintain computational efficiency~\cite{hu2019three}. However, accurately modeling turbulence remains a significant challenge~\cite{zou2020development}. Turbulence effects are implemented by replacing the molecular viscosity $\mu$ with an effective viscosity $\mu_{\text{eff}}$ that includes eddy viscosity $\mu^t$~\cite{hu2021sam}:

\begin{equation}
\mu_{\text{eff}} = \mu + \mu^t.
\label{eq:effective_viscosity}
\end{equation}

The eddy viscosity $\mu^t$ is computed using a simple zero-equation model based on Prandtl’s mixing-length theory~\cite{odier2009fluid}:

\begin{equation}
\mu^t = \rho L_{\text{mix}}^2 \sqrt{2 \sum_{i,j=1}^{3} S_{ij} S_{ij}},
\label{eq:turbulent_viscosity}
\end{equation}

where $\rho$ is the fluid density, $L_{\text{mix}}$ is the mixing length, and $S_{ij}$ is the strain rate tensor:

\begin{equation}
S_{ij} = \frac{1}{2} \left( \frac{\partial v_i}{\partial x_j} + \frac{\partial v_j}{\partial x_i} \right).
\label{eq:strain_tensor}
\end{equation}

This approach relies on turbulence models that use the eddy viscosity concept. However, algebraic turbulence models like Prandtl's mixing length have limited applicability and perform poorly in complex flow scenarios~\cite{huang1999physics}. Two-equation models, such as the $k$--$\epsilon$ model, still depend on the eddy viscosity concept but offer higher accuracy and better applicability.

To leverage the capabilities of higher-fidelity turbulence models within SAM, a data-driven closure surrogate model can be employed. In previous work, we introduced a Densely connected Convolutional Neural Network (DCNN) trained using high-fidelity CFD results from fine-mesh simulations~\cite{liu2022data,liu2022sam}. This DCNN acts as a closure model for eddy viscosity within SAM, enhancing its predictive capabilities for turbulent flows. The data-driven model aims to capture complex turbulence effects that are not adequately represented by traditional algebraic models.

\subsection{Uncertainty Quantification of DNN Models} 

While DNNs like the DCNN used in this work offer improved predictive capabilities, quantifying the uncertainties associated with these predictions is essential, especially in safety-critical applications like nuclear reactor modeling. Uncertainty in DNN models arises from various sources, including model parameters, data noise, and model structure. Providing reliable UQ allows for better risk assessment and decision-making.

Previous studies have explored different UQ methods for DNNs applied to turbulence modeling. In our earlier work, we evaluated three UQ methods: Monte Carlo (MC) dropout, DEs, and Bayesian Neural Networks (BNNs)~\cite{liu2021uncertainty_a}. Among these, the DE method showed the greatest potential in terms of computational efficiency, scalability, and predictive accuracy for UQ. DEs involve training multiple neural networks independently and combining their predictions to estimate uncertainty.

However, significant aleatoric uncertainty was observed, even with noise-free data, and discrepancies existed between DCNN predictions and CFD results. Aleatoric uncertainty, which arises from inherent noise in the data, was unexpectedly high, suggesting that the model might be misinterpreting epistemic uncertainty (model uncertainty) as aleatoric. This misinterpretation can lead to overconfidence in predictions and underestimation of model uncertainty.

Optimizing the neural network's hyperparameters is crucial to minimize these discrepancies and improve UQ. Hyperparameters such as the number of layers, learning rate, and activation functions significantly affect the model's performance and its ability to generalize~\cite{raiaan2024systematic}. Integrating optimization techniques, such as BO, with DNNs can significantly enhance both prediction accuracy and UQ~\cite{egele2022autodeuq,balaprakash2018deephyper}.

BO uses a surrogate model to guide the search for optimal hyperparameters, balancing exploration and exploitation in the hyperparameter space~\cite{tuba2021convolutional,dabiran2023sparse}. By systematically exploring the hyperparameter space, BO can find configurations that improve the model's performance more efficiently than random or grid searches.

\subsection{Motivation}

The motivation for this study arises from the need to enhance the performance of DNN models for eddy viscosity prediction in nuclear reactor safety analyses while providing reliable UQ. The existing DCNN model, although promising, exhibits significant aleatoric uncertainty and discrepancies with high-fidelity CFD data. These issues could undermine the reliability of the model in safety-critical applications.

By combining BO with DE—a method we refer to as BODE—we aim to improve the predictive accuracy of the DCNN and provide high-quality UQ. Unlike traditional DEs, where ensemble members are trained with the same hyperparameters, BODE optimizes hyperparameters for each ensemble member, potentially capturing a more diverse set of models and better representing epistemic uncertainty.

Our approach involves optimizing the DCNN's hyperparameters when forming the ensemble, rather than simply retraining the same network multiple times with different initializations. We assess BODE's performance by comparing it to a traditional baseline ensemble, where each ensemble member shares the same DCNN hyperparameters but is trained with different initial weights and data shuffling.

Furthermore, we analyze the capability of BODE to estimate uncertainties by incorporating Gaussian noise with standard deviations of 5\% and 10\% into the eddy viscosity data. This allows us to evaluate how well BODE captures data noise and distinguishes between aleatoric and epistemic uncertainties. Accurately estimating aleatoric uncertainty is crucial for understanding the inherent variability in the data, while epistemic uncertainty quantifies the confidence in the model itself.

The remainder of the paper is structured as follows: Section~\ref{sec:methodology} outlines the methodology, focusing on DE and UQ, the optimization strategy involving the Sobol sequence and BO implementation, and the integration of Gaussian noise. Section~\ref{sec:case_study} presents the case study of CFD eddy viscosity fields and details the DCNN employed for prediction. Section~\ref{sec:results} discusses the results of hyperparameter optimization, UQ of BODE in both noise-free and noisy environments. Finally, Section~\ref{sec:conclusions} concludes the paper and suggests directions for future research.

\section{Methodology}
\label{sec:methodology}

In this work, we propose a novel approach that combines BO with DE, referred to as BODE, to enhance both prediction accuracy and UQ in DNNs. Our methodology involves integrating BO into the process of forming a DE, aiming to optimize the hyperparameters of individual ensemble members to improve their performance and diversity.

\subsection{Deep Ensemble and Uncertainty Quantification}

DEs are a powerful method for improving the generalization and uncertainty estimation of neural networks~\cite{lakshminarayanan2017simple}. Instead of solely predicting the target value, DEs estimate predictive uncertainty by leveraging an ensemble of neural networks, each trained independently~\cite{YANG2024107871}. This approach not only enhances the model's generalization but also helps reduce overfitting.

In our approach, we select ensemble members using BO, choosing twenty models from twenty optimization processes to form the ensemble. The DE method assumes that the data follow a Gaussian distribution, where predictions are represented by a normal distribution characterized by both a mean and a variance. The estimated variance captures the aleatoric uncertainty, while the use of multiple models in the ensemble provides an estimate of the epistemic uncertainty~\cite{Laves2021RecalibrationOA}.

Aleatoric uncertainty refers to the inherent randomness and noise within the data. This type of uncertainty is irreducible, as it is a natural part of the data; the model can estimate it accurately but cannot reduce it. In contrast, epistemic uncertainty, also known as systematic uncertainty, arises from a lack of knowledge about the best model. Unlike aleatoric uncertainty, epistemic uncertainty can be reduced by collecting more data or refining the model~\cite{Hullermeier2021Aleatoric}.

We use the negative log-likelihood (NLL) of a Gaussian distribution as the loss function for training the neural networks. Assuming the data follow a Gaussian distribution, predictions are characterized by both a mean $\mu(x)$ and a variance $\sigma^2(x)$, as shown in Equation~\ref{eq:gaussian}.

\begin{equation}
y(x) \sim \mathcal{N}(\mu(x), \sigma^2(x))
\label{eq:gaussian}
\end{equation}

The probability of predicting a point $y_n$ given an input vector $x_n$ in a model parameterized by $\theta$ is expressed as:

\begin{equation}
p_\theta(y_n|x_n) = \frac{1}{\sqrt{2\pi\sigma^2(x_n)}} \exp \left( -\frac{(y_n - \mu(x_n))^2}{2\sigma^2(x_n)} \right)
\label{eq:NLL_p}
\end{equation}

Maximizing the likelihood is equivalent to minimizing the negative log-likelihood, which simplifies the computation for optimization:

\begin{equation}
- \log p_\theta(y_n|x_n) = \frac{\log \sigma^2(x_n)}{2} + \frac{(y_n - \mu(x_n))^2}{2\sigma^2(x_n)}
\label{eq:NLL}
\end{equation}

The loss function $C$ is defined as the sum of the negative log-likelihoods over all data points $n$:

\begin{equation}
C = \sum_n \left[ \frac{(y_n - \mu(x_n))^2}{2\sigma^2(x_n)} + \frac{\log \sigma^2(x_n)}{2} \right]
\label{eq:loss_nll}
\end{equation}

In this equation, $\sigma^2(x_n)$ represents the aleatoric variance, as it captures the data noise assuming a Gaussian distribution. Since epistemic uncertainty is related to the model's ability to generalize, we estimate it by training multiple models. An ensemble of $M$ models is trained, and the final prediction is obtained by averaging the outputs of these models:

\begin{equation}
\mu(x) = \frac{1}{M} \sum_{i=1}^M \mu_i(x)
\label{eq:avg_mean}
\end{equation}

The total uncertainty $\sigma^2(x)$ is calculated using the individual models' uncertainties $\sigma_i^2(x)$ and their predictions $\mu_i(x)$:

\begin{equation}
\sigma^2(x) = \frac{1}{M} \sum_{i=1}^M \left( \sigma_i^2(x) + \mu_i^2(x) \right) - \mu^2(x)
\label{eq:total}
\end{equation}

We can decompose the total uncertainty into aleatoric and epistemic uncertainties, as follows:

\begin{equation}
\sigma^2_{\text{aleatoric}}(x) = \frac{1}{M} \sum_{i=1}^M \sigma_i^2(x)
\label{eq:aleatoric}
\end{equation}

\begin{equation}
\sigma^2_{\text{epistemic}}(x) = \frac{1}{M} \sum_{i=1}^M \left( \mu_i^2(x) \right) - \mu^2(x)
\label{eq:epistemic}
\end{equation}

To assess the performance of our proposed BODE method, we utilize a baseline ensemble (BE) as a reference. The BE hyperparameters are determined through trial and error, where various configurations are tested manually, and the best-performing model is selected. In the BE, each model in the ensemble shares the same neural network architecture, but different weight initializations and data shuffling are applied using different random seeds.

\subsection{Optimization Strategy}

In our methodology, we employ the Ax-platform~\cite{Bakshy2018Ae, chang2019bayesian}, an open-source framework developed by Meta and built on PyTorch~\cite{paszke2019pytorch}, to perform the BO. The objective function is defined as the Root Mean Square Error (RMSE), as given in Equation~\ref{eq:rmse}, with the independent variables being the hyperparameters of the DCNN. Here, $\mu$ is the predicted mean from Equation~\ref{eq:avg_mean}.

\begin{equation}
\text{RMSE} = \sqrt{\frac{1}{n} \sum_{i=1}^{n} (y_i - \mu_i)^2}
\label{eq:rmse}
\end{equation}

While the negative log-likelihood (NLL) loss function is used during training for backpropagation, we optimize the RMSE because it provides a straightforward metric for evaluating prediction accuracy. Optimizing RMSE helps reduce the term $(y_n - \mu(x_n))^2$ in the NLL, thereby lowering the overall loss.

The optimization is conducted using 30\% of the data, which is split into optimization training and validation datasets with a 7:3 ratio. We minimize the RMSE on the optimization validation dataset to reduce the risk of overfitting.

Our optimization strategy involves two steps:

\begin{enumerate}
    \item \textbf{Sobol Sequence Initialization:} We use a Sobol sequence, a quasi-random sequence that produces points designed to fill a space uniformly, to generate samples covering the entire multidimensional hyperparameter search space~\cite{sun2021comparing}. This step builds the prior distribution of the Gaussian Process (GP) surrogate model.
    \item \textbf{Bayesian Optimization:} We employ BoTorch~\cite{Balandat2020BoTorch}, a programming framework for BO, to conduct the optimization. Based on the GP prior distribution built using the Sobol sequence, the acquisition function determines the next hyperparameter configuration to sample.
\end{enumerate}

During the Sobol sequence step, we generate $N_0$ hyperparameter setups. After completing the Sobol sequence observations, we proceed with BO for $N_A$ iterations, making the total number of iterations $N_T = N_0 + N_A$. The entire optimization algorithm is outlined in Algorithm~\ref{alg:bo}.

We optimize eight hyperparameters, including the learning rate, weight decay, drop rate, batch size, number of dense blocks, number of layers within each dense block, growth rate, and size of initial features. Each optimization model is initialized with different values for these hyperparameters. The hyperparameter configurations obtained from BO are then used to train the models, and the optimized models are combined using DE.

\begin{algorithm}
\caption{Optimization Strategy (where $M$ represents the model index and $i$ represents the iteration number).}
\label{alg:bo}
\begin{algorithmic}[1]
\For{$M = 1$ to $M_{ensemble}$}
    \State Choose a random hyperparameter setup from the search space to initialize the Sobol sequence
    \For{$i = 1$ to $N_0$}
        \State Apply the Sobol sequence to uniformly explore the hyperparameter search space
        \State Train the DCNN using the current hyperparameter setup
        \State Evaluate the RMSE on the optimization validation dataset
        \State Update the prior distribution of the GP surrogate model
    \EndFor
    \For{$i = N_0 + 1$ to $N_T$}
        \State Use the acquisition function to sample a new hyperparameter setup $x_{n+1}$
        \State Train the model with the new setup and evaluate its RMSE on the optimization validation dataset
        \State Update the GP surrogate model with the new observation
    \EndFor
\EndFor
\State Identify the hyperparameter setup that results in the lowest RMSE (best-performing model)
\end{algorithmic}
\end{algorithm}

\subsection{Sobol Sequence}

To initialize the GP surrogate model, we use the Sobol sequence to generate quasi-random samples that uniformly cover the hyperparameter search space~\cite{sun2021comparing}. The Sobol sequence uses a set of direction numbers \(V_{d, m}\) for each dimension \(d\), where \(m\) ranges from 1 to \texttt{MAXBIT}, as given in Equation~\ref{eq:direc}.

\begin{equation}
V_{d, 1}, V_{d, 2}, \dots, V_{d, m}
\label{eq:direc}
\end{equation}

For \(j > m\), the direction numbers are recursively computed as:

\begin{equation}
V_{d, j} = V_{d, j-m} \oplus \left( \sum_{k=1}^{m} c_k \cdot 2^{k} \cdot V_{d, j-k} \right)
\label{eq:recursive}
\end{equation}

where \(V_{d, j-m}\) is the previously computed direction number, \(c_k\) are binary coefficients (0 or 1), and \(\oplus\) denotes the XOR operation. The direction numbers are scaled by a power of 2 to transform the binary fraction into fixed-point numbers:

\begin{equation}
V_{d, j} = V_{d, j} \times 2^{\text{(MAXBIT - j)}}
\label{eq:power2}
\end{equation}

The \(n\)-th point \(\mathbf{x}_n = (x_{n, 1}, x_{n, 2}, \dots, x_{n, d})\) in the sequence is generated as:

\begin{equation}
x_{n, d} = \frac{1}{2^{b}} \left( \sum_{m=1}^{b} b_m \cdot V_{d, m} \right)
\label{eq:seq}
\end{equation}

where \(b_m\) are the binary digits of the integer \(n\), producing a quasi-random sample in the unit interval \([0, 1]\).

\subsection{Bayesian Optimization}

BO is an effective technique for optimizing black-box functions that are costly to evaluate~\cite{Snoek2012Practical}. The goal is to identify the set of input parameters that result in the minimum (or maximum) value of the objective function.

BO consists of two key components: the surrogate model and the acquisition function. The surrogate model is a probabilistic model that approximates the objective function based on a limited number of evaluations. As more points are sampled, the surrogate model becomes more accurate. We use a Gaussian Process (GP) as the surrogate model, specifically the \textit{SingleTaskGP} from BoTorch~\cite{Balandat2020BoTorch}.

A GP is defined by a mean function \( \mu(x) \) and a covariance function \( k(x,x') \). The RMSE is approximated using the GP as shown in Equation~\ref{eq:gp}.

\begin{equation}
\text{RMSE} \sim \mathcal{GP}(\mu(x), k(x,x'))
\label{eq:gp}
\end{equation}

In our work, we use the Radial Basis Function (RBF) as the covariance function:

\begin{equation}
k_{\text{RBF}}(x,x') = \exp\left( -\frac{\|x - x'\|^2}{2\ell^2} \right)
\label{eq:rbf}
\end{equation}

The BO process iteratively updates the surrogate model using the acquisition function, which determines the next point to evaluate. We use the \textit{qNoisyExpectedImprovement} acquisition function~\cite{Balandat2020BoTorch}, which computes the expected improvement over the best observed value:

\begin{equation}
q\text{NEI}(\mathcal{X}) = \mathbb{E} \left[ \max \left( \min(Y) - \min (Y_{\text{baseline}}), 0 \right) \right]
\label{eq:acquisition}
\end{equation}

By integrating BO with DE, we aim to improve both prediction accuracy and uncertainty estimation, particularly in the presence of data noise. Minimizing RMSE helps reduce the prediction errors, leading to lower epistemic uncertainty as the variance between ensemble members' predictions decreases.

During training, we perform gradient backpropagation using the negative log-likelihood loss function. By optimizing RMSE and using the NLL loss, the model becomes better at capturing noise in the data, reducing aleatoric uncertainty, and delivering more accurate predictions. This approach results in a model that not only fits the data well but also provides reliable uncertainty estimates.

\subsection{Gaussian Noise}

Since the training data are derived from numerical simulations, they are noise-free. To verify the validity of the BODE method in quantifying uncertainties, we introduce Gaussian noise into the data. We consider two cases with standard deviations of 5\% and 10\%.

We sample a noise factor $\epsilon$ from a Gaussian distribution with mean zero and a specified standard deviation:

\begin{equation}
\epsilon = \mathcal{N}(0, \sigma^2)
\label{eq:noise_factor}
\end{equation}

The noise factor is then smoothed using a Gaussian filter:

\begin{equation}
\mathcal{N} = \frac{1}{\sqrt{2\pi\sigma^2}} \exp\left(-\frac{x^2}{2\sigma^2}\right)
\label{eq:gauss_noise}
\end{equation}

\begin{equation}
\tilde{\epsilon}(x) = (\epsilon * \mathcal{N})(x)
\label{eq:filter}
\end{equation}

To generate noise that follows the pattern and is proportional to the CFD results, we multiply the noise factor by the original eddy viscosity $\mu^t$ values. The noisy data are then obtained as:

\begin{equation}
y = \max(0, \mu^t(x) + \tilde{\epsilon}(x) \cdot \mu^t(x))
\label{eq:final_noise}
\end{equation}

This process is repeated for each timestep at every epoch, reinforcing the Gaussian assumption since the noise is applied consistently to each data cell across all epochs.

\section{Case Study}
\label{sec:case_study}

In this case study, we apply the proposed BODE method to predict the eddy viscosity field in a nuclear reactor thermal stratification scenario.

\subsection{Computational Fluid Dynamics Database}

\begin{figure}[H]
    \centering
    \includegraphics[width=0.5\linewidth]{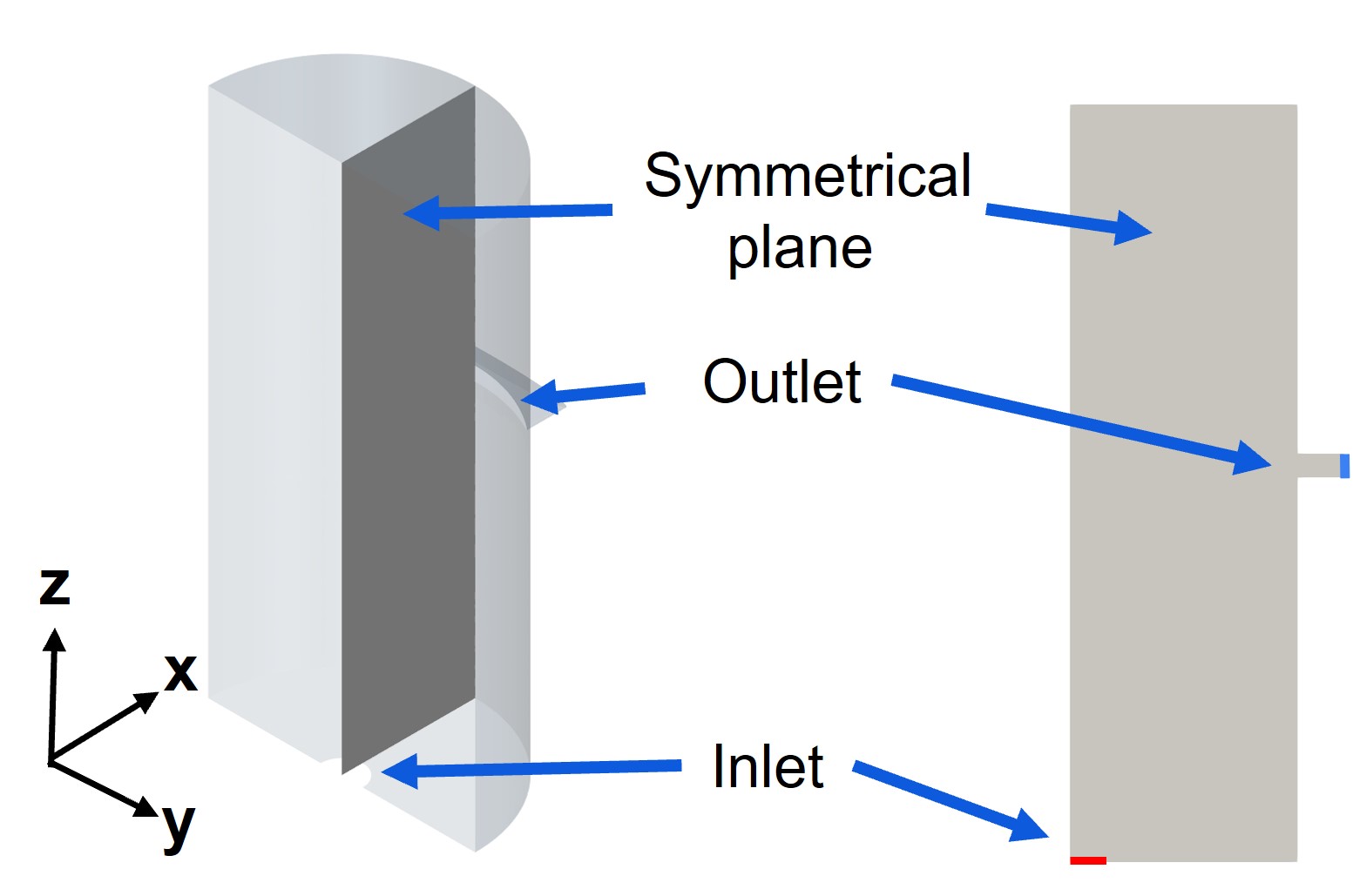}
    \caption{Computational domain for the CFD simulations: half of the SFR hot pool cylindrical tank (left), the symmetry plane (right).}
    \label{fig:cfd}
\end{figure}

We use CFD results for the fluid fields, as reported in previous work by Liu et al.~\cite{liu2022data}. The CFD model represents a cylindrical tank simulating the hot pool of a Sodium-cooled Fast Reactor (SFR), with dimensions of 7.74~m in height and 1.95~m in radius. The turbulence model applied is the Realizable $k$--$\epsilon$ formulation within the Reynolds-Averaged Navier-Stokes (RANS) framework. To simplify the geometry, only half of the cylinder is modeled using a symmetry boundary condition. Figure~\ref{fig:cfd} depicts the computational domain used for the CFD simulations.

The transient under study is a loss-of-flow (LOF) event, which assumes a complete loss of forced circulation due to multiple system failures. In such a case, heat removal relies entirely on natural convection. The transient is defined according to the conceptual design report for the Advanced Burner Test Reactor (ABTR), a pool-type SFR~\cite{abtr}.

The LOF scenario for the ABTR was first analyzed using the 1-D module of the System Analysis Module (SAM). These 1-D simulation results were then used to define the inlet boundary conditions for the CFD model. The CFD transient was solved using STAR-CCM+~\cite{siemens2018simcenter}, with time-varying inlet temperature and mass flow rate inputs. The initial temperature and mass flow rate were set to 783.15~K and 632~kg/s, respectively~\cite{liu2021uncertainty_a}.

For the DCNN analysis, the X-Z plane shown in Figure~\ref{fig:cfd} was selected, with the red line indicating the area designated for quantitative analysis. The plane extends from the inlet to the outlet and is centered in the middle of the tank. The transient simulations were run for 600~s. To ensure that the time-sequence data have equal weight in training the model, the simulation results were extracted with a uniform time step of 0.2~s, resulting in 3000 timesteps.

In the CFD simulation, the original setup uses an unstructured mesh with more than one million cells. Since these cells have varying sizes and shapes, they cannot be directly used to train the DCNN, which assumes each input cell has the same significance. Additionally, this nonuniform mesh is incompatible with the coarse mesh setup required by the system code. To address this, we use the k-Nearest Neighbor (kNN) algorithm to convert the unstructured mesh into a structured, uniform mesh with dimensions of $32 \times 64 \times 128$, making it suitable for training and integration with the system.

\subsection{Densely Connected Convolutional Neural Network}

\begin{figure}[H]
    \centering
    \includegraphics[width=\linewidth]{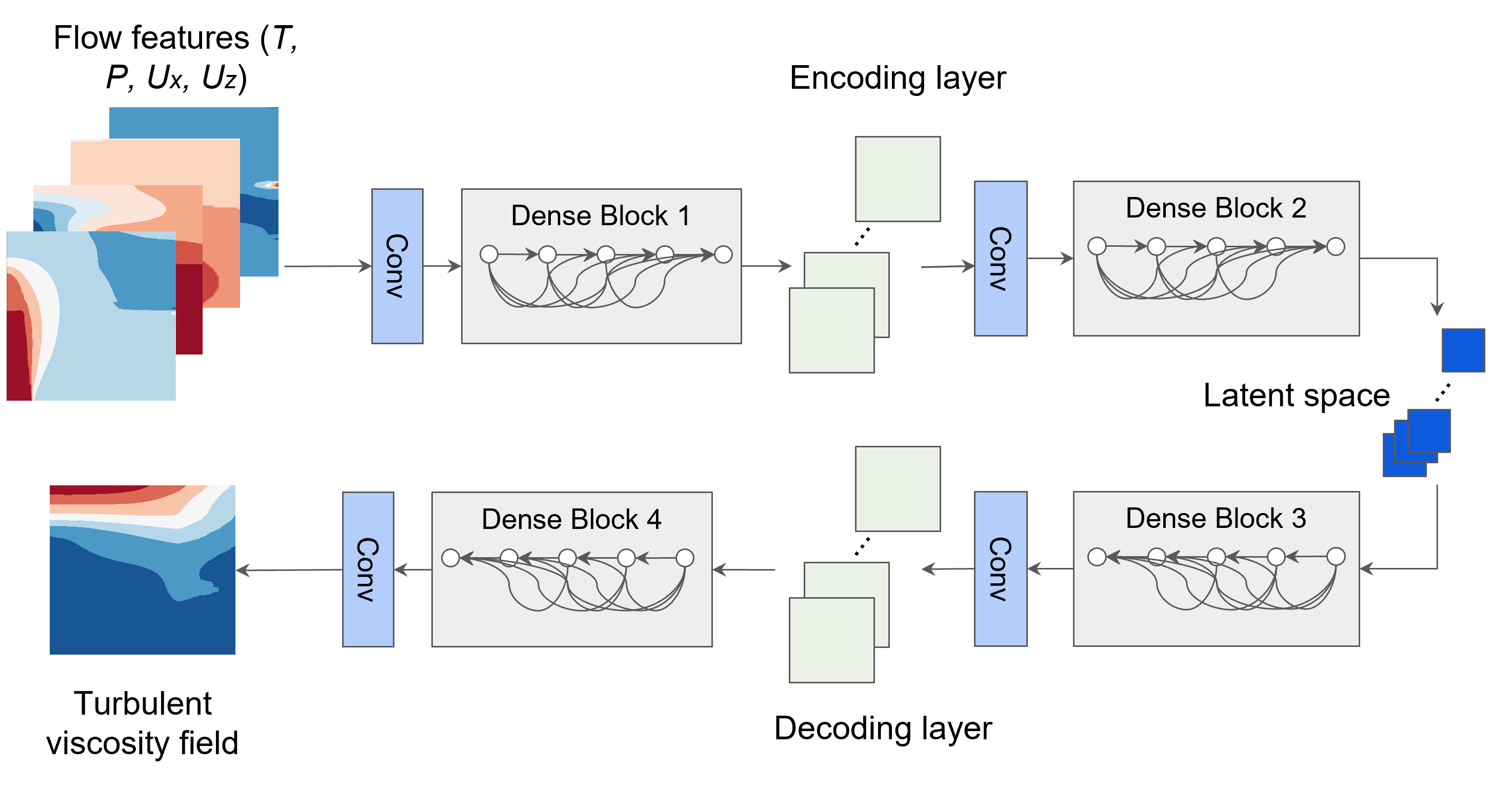}
    \caption{Architecture of the Densely Connected Convolutional Neural Network.}
    \label{fig:dcnn}
\end{figure}

We employ a DCNN for predicting the eddy viscosity field. A DCNN is an enhanced version of a Convolutional Neural Network (CNN), composed of several dense blocks of convolutional layers, where each layer is directly connected to all subsequent layers within the block~\cite{huang2019convolutional}. The output of each convolution is combined with the original input along the channel axis. This architecture enhances the flow of information during forward propagation and improves gradient flow during backpropagation, mitigating the vanishing gradient problem and ensuring better feature propagation. DCNN has demonstrated good performance in field-to-field prediction of thermal-fluid problems~\cite{zhu2018bayesian, liu2022data}. The DCNN architecture is illustrated in Figure~\ref{fig:dcnn}.

The DCNN model is designed to learn from five input features, including temperature ($T$), pressure ($p$), and the three velocity components ($U_x$, $U_y$, $U_z$), to predict the eddy viscosity ($\mu^t$). The DCNN follows a standard auto-encoder design and consists of an odd number of dense blocks. The model utilizes several convolutional layers and applies Rectified Linear Unit (ReLU) activation functions. To guarantee non-negative predictions, the output is activated using a Softplus function. Since the analysis is conducted in two dimensions, one velocity component is omitted.

A DCNN configuration involves several hyperparameters, including the learning rate, which determines the adjustment of weight coefficients during gradient descent, and weight decay, a regularization parameter that penalizes large weight values to prevent overfitting. Other important hyperparameters include the drop rate, defining the probability of dropping units during training to reduce overfitting; the batch size, which specifies the number of training samples processed in each iteration; the number of dense blocks; the number of layers within each dense block; the growth rate, which controls the number of output feature maps added by each convolutional layer~\cite{huang2019convolutional}; and the initial size of the features, representing the number of filters in the first convolutional layer.

For the baseline ensemble, we determined the hyperparameters through manual tuning and trial-and-error experiments~\cite{liu2021uncertainty_a}. Various configurations were tested by adjusting one hyperparameter at a time while keeping others fixed, to understand their individual impact on model performance. The final hyperparameters for the baseline ensemble were selected based on the best observed performance on the validation dataset. The final baseline ensemble hyperparameters are shown in Table~\ref{table:baseline_hyper}. In this baseline ensemble, each model shares the same neural network architecture, but different weight initializations and data shuffling are applied using different random seeds.

\begin{table}[H]
    \centering
    \caption{Baseline ensemble hyperparameters.}
    \begin{tabular}{ll}
        \toprule
        \textbf{Hyperparameter}     & \textbf{Value}      \\
        \midrule
        Batch size                  & 16                  \\
        Learning rate               & 0.0008              \\
        Weight decay                & 0.002               \\
        Number of dense blocks      & 3                   \\
        Layers in dense blocks      & 3, 4, 5             \\
        Growth rate                 & 16                  \\
        Drop rate                   & 0.15                \\
        Initial features            & 32                  \\
        Output activation function  & Softplus            \\
        Inner layers activation function & ReLU           \\
        Epochs                      & 200                 \\
        \bottomrule
    \end{tabular}
    \label{table:baseline_hyper}
\end{table}

For the BODE models, we optimize the hyperparameters within specified ranges or sets, as summarized in Table~\ref{tab:hyperparameters}. The selection of these ranges is based on prior experience and literature, aiming to balance model complexity and computational efficiency.

\begin{itemize}
    \item \textbf{Learning rate}: We consider a range from $10^{-4}$ to $10^{-2}$, covering two orders of magnitude, to explore both conservative and aggressive learning rates that influence the speed and stability of convergence.
    \item \textbf{Weight decay}: Similar to the learning rate, we use a range from $10^{-4}$ to $10^{-2}$ to examine the effect of different levels of regularization on model generalization.
    \item \textbf{Drop rate}: The range from 0 to 0.5 allows us to assess the impact of dropout regularization from no dropout to substantial dropout, which can prevent overfitting but may also hinder learning if too high.
    \item \textbf{Number of dense blocks}: We choose values of 3 and 5 to investigate architectures with different depths, balancing between model capacity and computational cost.
    \item \textbf{Number of layers within each dense block}: A range from 3 to 9 layers enables us to explore varying levels of feature extraction complexity within each block.
    \item \textbf{Batch size}: The set includes values from 8 to 128 to assess how different batch sizes affect training stability and convergence, considering computational resource limitations.
    \item \textbf{Growth rate}: Values from 8 to 48 are selected to explore how rapidly the number of feature maps increases in the network, affecting model capacity and computational requirements.
    \item \textbf{Initial features}: A range from 8 to 128 allows us to examine the impact of the initial number of filters, influencing the model's ability to capture low-level features.
\end{itemize}

By optimizing over these ranges, we aim to identify hyperparameter configurations that improve model performance while avoiding overfitting. The inclusion of a broad range of values allows the BO process to explore diverse architectures and training dynamics, potentially leading to models with better generalization and uncertainty estimation capabilities.
\begin{table}[H]
\centering
\caption{Hyperparameter search space for BODE models.}
\begin{tabular}{@{}ll@{}}
\toprule
\textbf{Hyperparameter} & \textbf{Range/Values} \\ \midrule
Learning rate           & [0.0001 – 0.01]       \\
Weight decay            & [0.0001 – 0.01]       \\
Drop rate               & [0 – 0.5]             \\
Number of dense blocks  & \{3, 5\}              \\
Number of layers        & [3–9]                 \\
Batch size              & \{8, 12, 16, 24, 32, 48, 64, 128\} \\
Growth rate             & \{8, 12, 16, 24, 32, 48\} \\
Initial features        & \{8, 12, 16, 24, 32, 48, 64, 128\} \\ \bottomrule
\end{tabular}
\label{tab:hyperparameters}
\end{table}

\subsection{Data Preprocessing}

Neural networks are sensitive to the scale and distribution of input features, often favoring larger values. In our case, the input features—pressure, temperature, and velocity—have different orders of magnitude. To address this, we normalize the data using z-normalization, as shown in Equation~\ref{eq:z}, where $x$ represents the input features, $\mu$ is the mean, and $\sigma$ is the standard deviation of the input feature.

\begin{equation}
z = \frac{x - \mu}{\sigma}
\label{eq:z}
\end{equation}

The dataset is divided into three groups: training, validation, and testing. The training set comprises 70\% of the total data, the validation set accounts for 29.5\%, and the remaining 0.5\% is reserved for testing. The testing dataset remains unseen during both the training and optimization phases, serving as a blind dataset. All reported results are based on the testing data.

For the optimization process, we apply BO to a subset containing 30\% of the combined training and validation data. Performing the optimization on the full dataset would be computationally expensive, particularly since the optimization is carried out over 75 iterations for each of the 20 models. Using a representative sample is sufficient for the optimization process.

\section{Results and Discussion}
\label{sec:results}
\subsection{Hyperparameter Optimization}

In this section, we present the results of hyperparameter optimization using BO. To reduce computational cost while maintaining sufficient data for effective optimization, we utilize 30\% of the available data sampled from both the training and validation datasets. This subset consists of 900 time steps, of which 70\% (630 time steps) are allocated for training (the BO training dataset), and the remaining 30\% (270 time steps) are designated for validation (the BO validation dataset).

The primary objective of the optimization is to minimize the Root Mean Square Error (RMSE) on the BO validation dataset. By focusing on the validation dataset rather than the training dataset, we aim to reduce the risk of overfitting.

The optimization process comprises 75 iterations, a number chosen heuristically. Initially, for the first 24 iterations, a Sobol sequence is used to establish a prior distribution for the Gaussian Process (GP) surrogate model. Subsequently, BO is implemented using BoTorch to identify the optimal hyperparameters. In each iteration, the DCNN model is trained on the BO training dataset and evaluated on the BO validation dataset based on the RMSE. These RMSE values serve as inputs to build and update the GP surrogate model. During each step of the BO process, the acquisition function selects the next hyperparameter configuration to sample based on the current state of the surrogate model.

\begin{figure}[H]  
    \centering
    \includegraphics[width=0.7\linewidth]{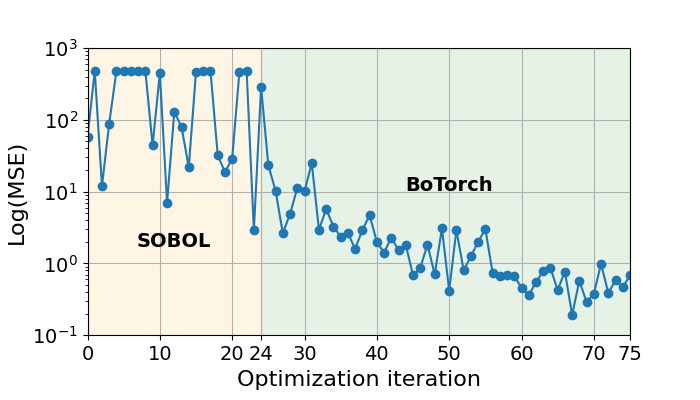} 
    \caption{The optimization algorathim iterations in Ax-platform.}
    \label{fig:bo_iterations}
\end{figure}

Figure~\ref{fig:bo_iterations} illustrates the RMSE trend across the optimization iterations. The RMSE decreases over the iterations, indicating that the surrogate model progressively captures the relationship between hyperparameters and the RMSE objective function more accurately with each additional observation. The fluctuations observed are due to the exploration aspect of the acquisition function, which balances exploration and exploitation during optimization. The best-performing model, identified by its optimized hyperparameter configuration after 75 iterations, is selected as an ensemble member.

To construct the ensemble, we repeat this optimization process twenty times in parallel, each initialized with a distinct set of hyperparameters generated by the Sobol sequence. This approach results in different prior Gaussian distributions in the GP surrogate models, ensuring diversity among the ensemble members at the end of the optimization.

\begin{figure}[H]  
    \centering
    \includegraphics[width=0.7\linewidth]{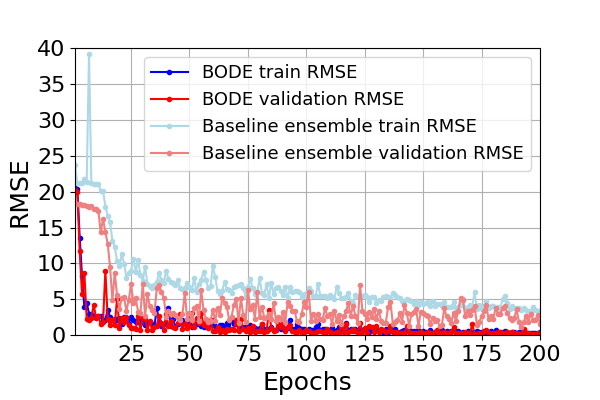} 
    \caption{The RMSE of the train and validation dataset for the best performing ensemble member of BODE and baseline ensemble.}
    \label{fig:loss}
\end{figure}

Following the optimization, the twenty selected hyperparameter configurations are used to train models on the full DCNN training dataset, which constitutes 70\% of the data. Once trained, these models are validated using the validation dataset, covering 29.5\% of the data. The training spans a total of 200 epochs. The training and validation RMSE for the best-performing models from both the BODE and the baseline ensemble are shown in Figure~\ref{fig:loss}. The RMSE decreases with each epoch in both the baseline ensemble and BODE models, with both models converging after approximately 150 epochs.

\begin{figure}[H]  
    \centering
    \includegraphics[width=1\linewidth]{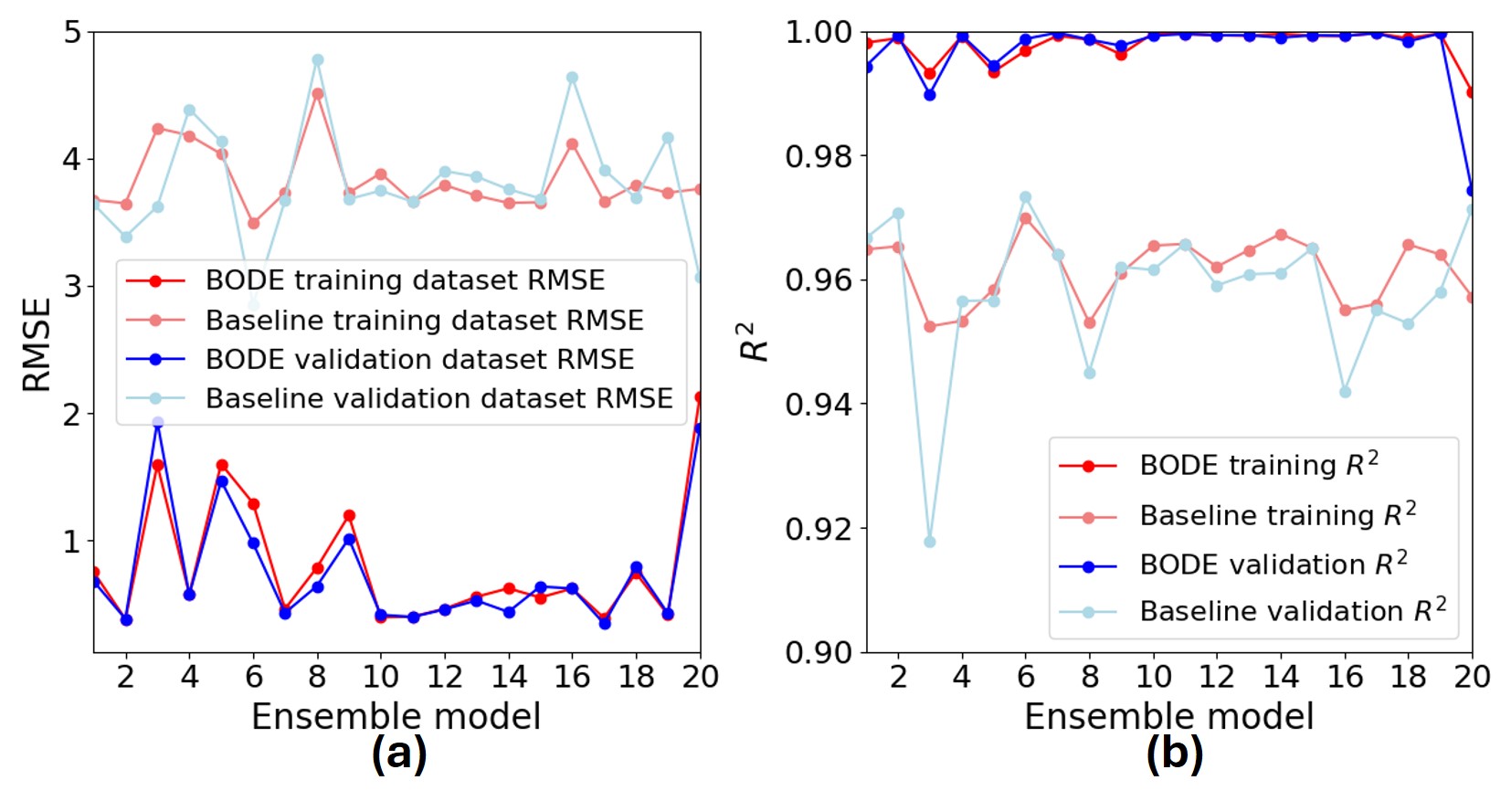}  
    \caption{Bayesian optimization results (a) in terms of RMSE (a) in comparison to the baseline ensemble (b) for the optimization iterations.}
    \label{fig:RMSE_R2}
\end{figure}

To evaluate and compare the performance of the BODE and baseline ensemble models, we use the RMSE and $R^2$ metrics, as shown in Figure~\ref{fig:RMSE_R2}. The RMSE values for the twenty ensemble models from both the BODE and baseline ensemble across the training and validation datasets are presented in Figure~\ref{fig:RMSE_R2}(a), while the corresponding $R^2$ values are shown in Figure~\ref{fig:RMSE_R2}(b).

The average RMSE for the BODE models on the training and validation datasets are 0.75~Pa$\cdot$s and 0.80~Pa$\cdot$s, respectively, whereas the baseline ensemble models exhibit higher average RMSE values of 3.81~Pa$\cdot$s and 3.83~Pa$\cdot$s. Regarding the $R^2$ metric, BODE demonstrates superior performance, with average $R^2$ values of 0.998 and 0.997 for the training and validation datasets, respectively. In contrast, the baseline ensemble models show lower average $R^2$ values of 0.962 and 0.958 for the same datasets.

It is noteworthy that the BODE validation RMSE is generally lower than the BODE training RMSE across most ensemble members. This observation is consistent with the fact that the optimization targets the validation dataset, and thus the models are better tuned for validation performance. Conversely, in the baseline ensemble models, the training RMSE is lower than the validation RMSE, indicating a potential overfitting to the training data.

\begin{figure}[H]  
    \centering
    \includegraphics[width=0.85\linewidth]{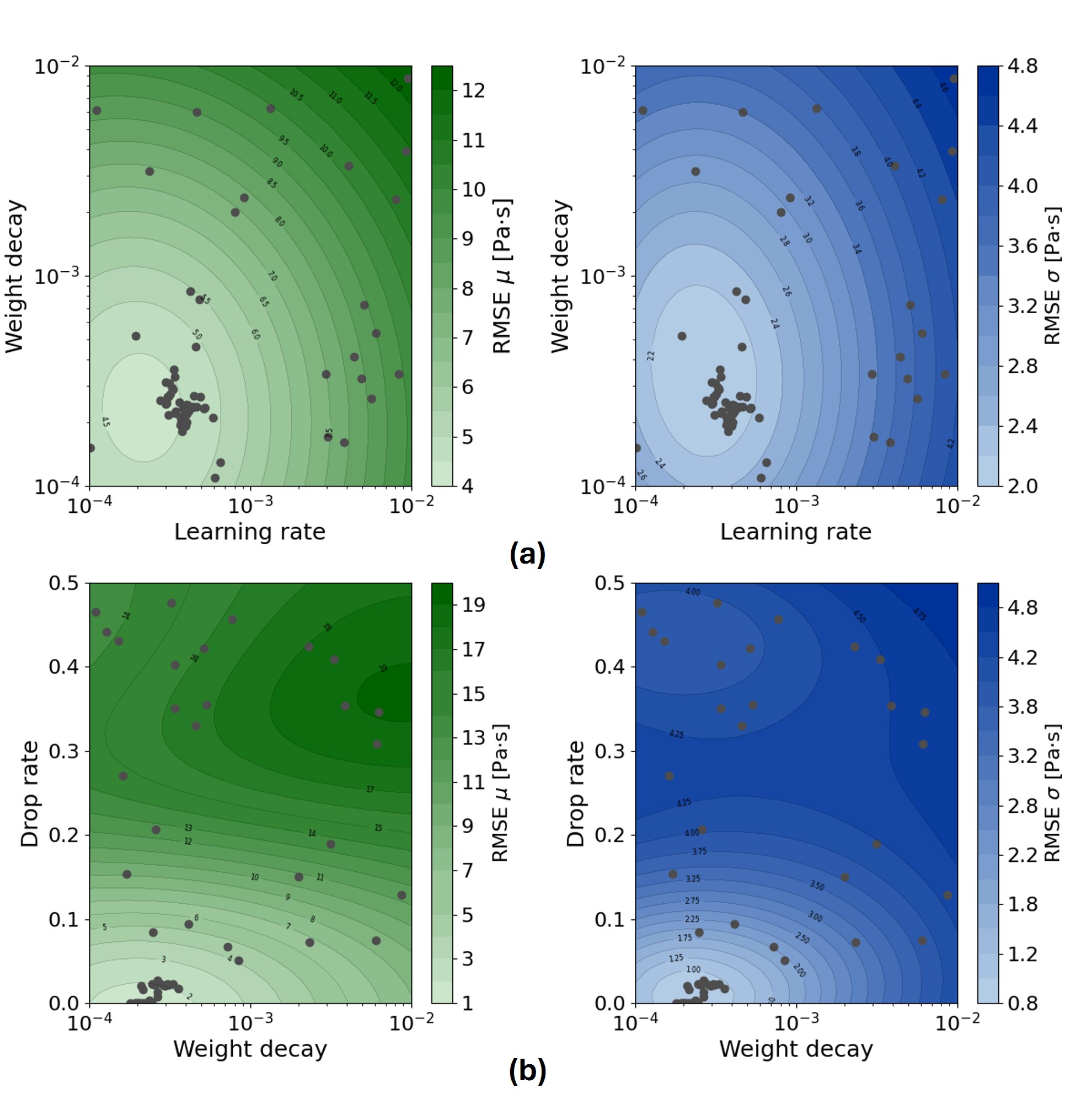} 
    \caption{The BO surrogate model results in terms of mean (left) and standard deviation (right) of the RMSE for (a) weight decay and learning rate, (b) drop rate and weight decay.}
    \label{fig:bo_hyperparameters}
\end{figure}

The BO process for the twenty ensemble models suggests that optimal values for the learning rate and weight decay, which yield the lowest RMSE, fall within the range of $10^{-4}$ to $10^{-3}$. Additionally, a low dropout rate, close to zero, enhances predictions and reduces RMSE, as shown in Figure~\ref{fig:bo_hyperparameters}.

Figure~\ref{fig:bo_hyperparameters}(a) displays contour plots of the GP surrogate model, representing the mean and standard deviation of the RMSE for the learning rate and weight decay hyperparameters, with scattered points indicating the observed BO iterations. Similarly, Figure~\ref{fig:bo_hyperparameters}(b) presents the mean and standard deviation of the RMSE for weight decay and dropout rate. The RMSE range in Figure~\ref{fig:bo_hyperparameters}(b) spans from 1~Pa$\cdot$s to 19~Pa$\cdot$s, while Figure~\ref{fig:bo_hyperparameters}(a) spans from 4~Pa$\cdot$s to 12~Pa$\cdot$s, highlighting that fine-tuning the dropout rate is crucial for further reducing the RMSE.

As previously noted, the dropout rate in the baseline ensemble models is set to 0.15, while considerably lower values are used in the BODE ensemble models, as determined by the optimization process. Since this optimization is conducted on the validation dataset, the risk of overfitting is minimized, validating the use of a lower dropout rate. Through manual iterations, the learning rate and weight decay used in the baseline ensemble are chosen to be 0.0008 and 0.002, respectively. The GP surrogate model identifies better performance when using a smaller value for the weight decay.

The BO process for the twenty ensemble models did not reveal a strong preference for other hyperparameters, such as growth rate, initial features, the number of dense blocks, or the number of layers within each dense block. The number of BO iterations (75) was determined by comparing the surrogate model's predictions with the observed RMSE after each iteration. Based on experimentation, a heuristic guideline suggests that the optimal number of iterations is approximately nine to ten times the number of hyperparameters, with around three times the number of hyperparameters allocated to the Sobol sequence for initialization.

\subsection{Noise-Free Environment Results}
In this section, we evaluate the performance of the BODE and baseline ensemble models in a noise-free environment. All the time step results presented are part of the testing dataset, which was not included in the training or optimization process.

\begin{figure}[H]  
    \centering
    \includegraphics[width=0.85\linewidth]{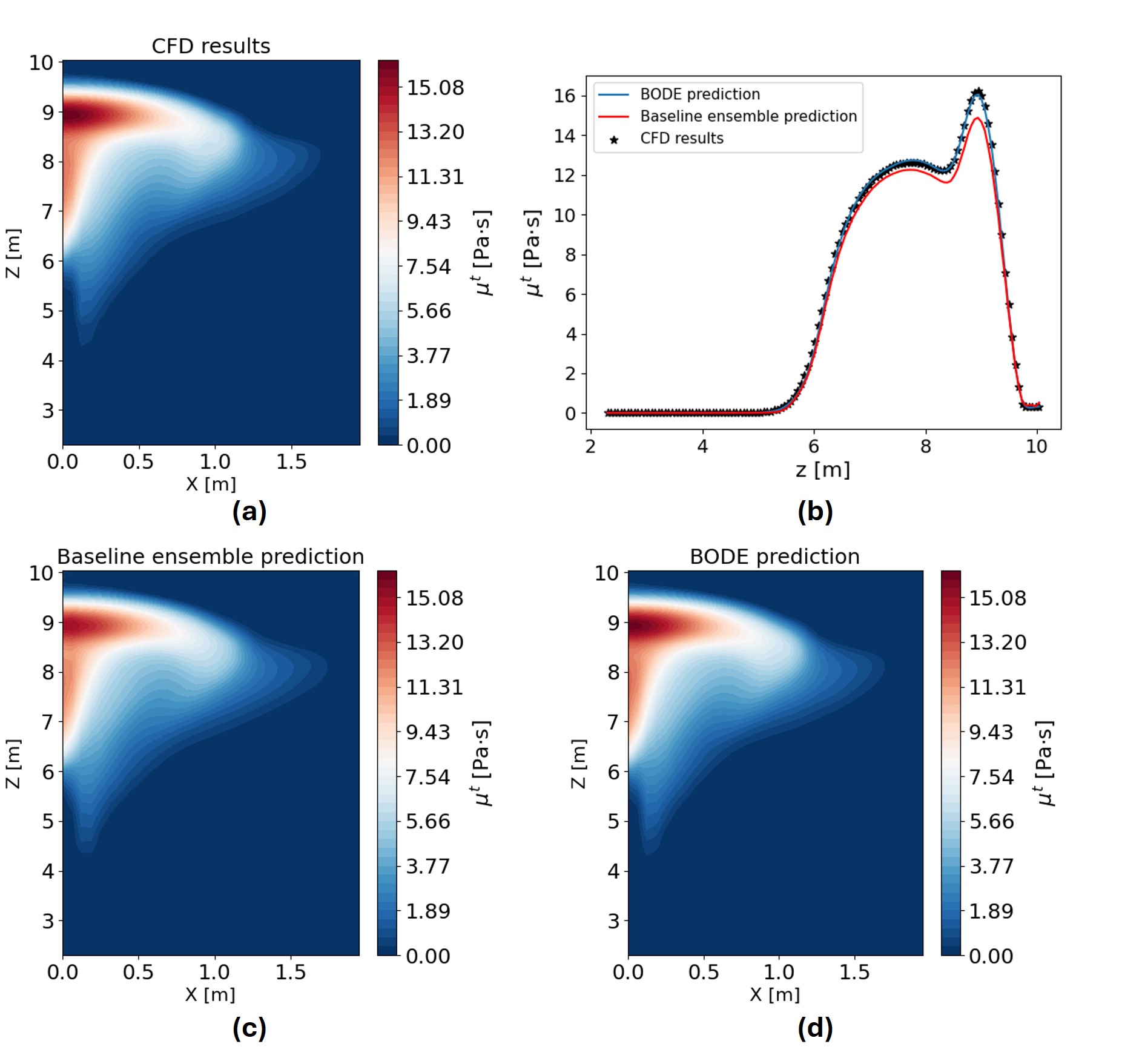}  
    \caption{Results at $t = 500$ $s$ : (a) CFD results for the eddy viscosity $\mu^t$; (b) Comparison of baseline ensemble prediction and BODE prediction over the symmetry line; (c) Baseline ensemble prediction of eddy viscosity $\mu^t$; and (d) BODE prediction of eddy viscosity $\mu^t$}
    \label{fig:prediction_2500}
\end{figure}

A comparison between the baseline ensemble and BODE predictions at $t = 500$~s is shown in Figure~\ref{fig:prediction_2500}. The results are presented in two formats: 2D spatial plots and a comparison along the symmetry line. Figure~\ref{fig:prediction_2500}(a) illustrates the original CFD results for the eddy viscosity $\mu^t$ (in Pa$\cdot$s). The plot displays the horizontal distance $X$ (in meters) and vertical distance $z$ (in meters), with the colormap representing the magnitude of the eddy viscosity $\mu^t$. At this time step, a region of higher viscosity appears near the upper left corner of the contour plot, centered between 8~m and 9~m vertically and extending horizontally to approximately 0.5~m. The eddy viscosity $\mu^t$ gradually decreases as the distance from this region increases.

The predictions from the baseline ensemble and BODE models are shown in Figures~\ref{fig:prediction_2500}(c) and \ref{fig:prediction_2500}(d), respectively. Both models effectively capture the overall features and patterns observed in the CFD results; however, the BODE model demonstrates higher accuracy, particularly in regions with elevated eddy viscosity values. This improvement is further confirmed in the quantitative comparison along the symmetry line, as shown in Figure~\ref{fig:prediction_2500}(b), where the BODE prediction exhibits a closer match to the CFD results.

\begin{figure}[H]  
    \centering
    \includegraphics[width=0.85\linewidth]{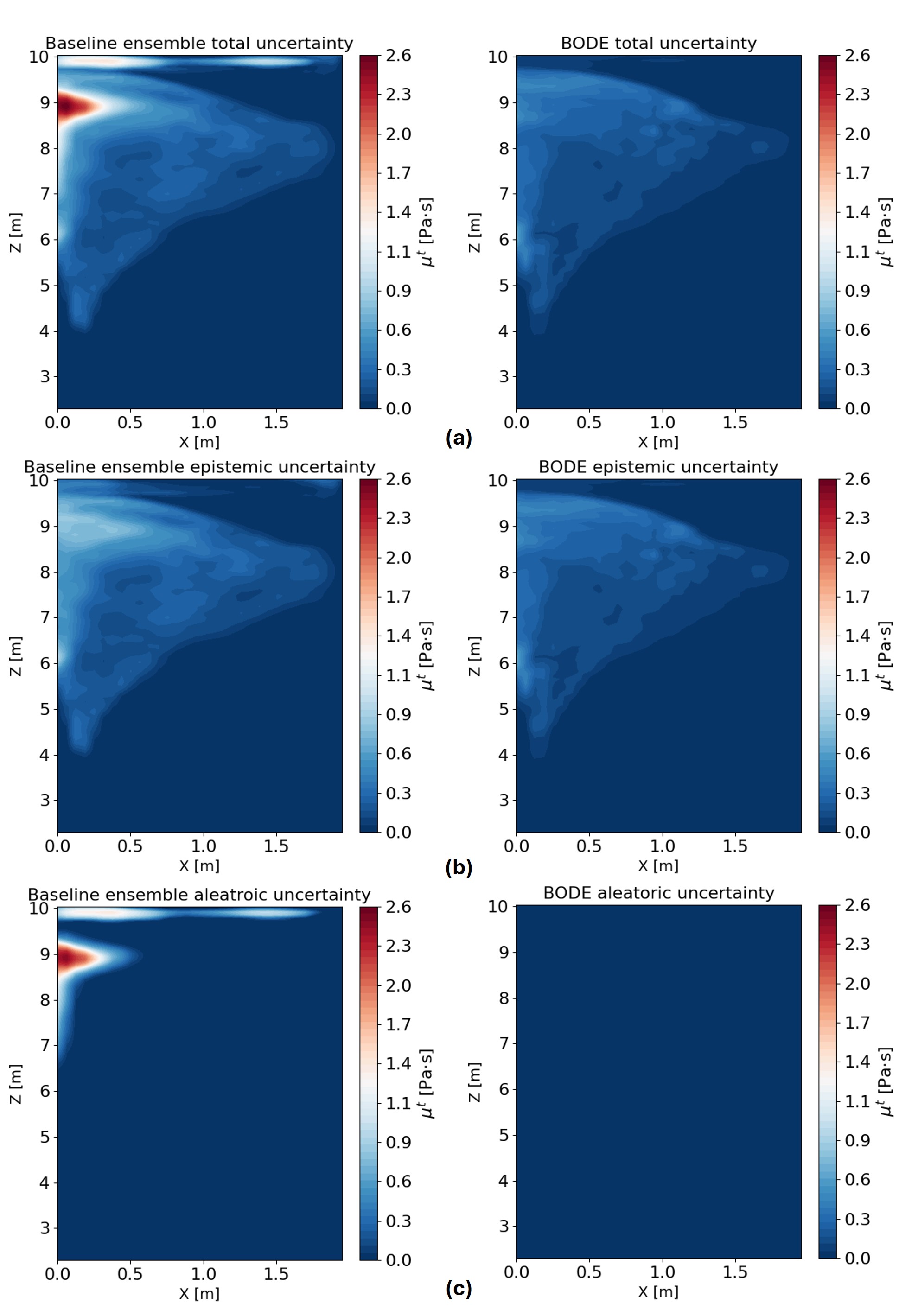}  
    \caption{Comparison of Baseline Ensemble and BODE Uncertainty: (a) Total uncertainty for the Baseline ensemble (left) and BODE (right); (b) Epistemic uncertainty for the Baseline ensemble (left) and BODE (right); (c) Aleatoric uncertainty for the Baseline ensemble (left) and BODE (right).}
    \label{fig:uncertainty_2500}
\end{figure}

The UQ results for both the baseline ensemble and the BODE at $t = 500$~s are presented in Figure~\ref{fig:uncertainty_2500}. The uncertainties are divided into three components: total uncertainty, epistemic uncertainty, and aleatoric uncertainty, as defined by Equations~\ref{eq:total}, \ref{eq:epistemic}, and \ref{eq:aleatoric}, respectively.

It can be observed that the baseline ensemble model exhibits a higher overall level of uncertainty compared to the BODE model. In the baseline ensemble, aleatoric uncertainty is the dominant component, significantly exceeding the epistemic uncertainty. In contrast, the BODE model predicts near-zero aleatoric uncertainty, suggesting that the remaining uncertainty is primarily epistemic. At higher eddy viscosity $\mu^t$ values, the baseline ensemble model predicts more than four times the total uncertainty estimated by the BODE, as shown in Figure~\ref{fig:uncertainty_2500}(a). The maximum total uncertainty estimated by the baseline ensemble reaches approximately 2.6~Pa$\cdot$s, while the BODE's maximum total uncertainty is only 0.6~Pa$\cdot$s. This comparison highlights the greater confidence in BODE predictions, as the baseline ensemble exhibits a broader distribution of high uncertainty levels.

The epistemic uncertainty, which reflects model knowledge gaps rather than data noise, tends to be more spread out across larger regions than the aleatoric uncertainty. The aleatoric uncertainty, which arises from inherent data noise, remains more localized. The baseline ensemble model estimates a maximum epistemic uncertainty of approximately 0.9~Pa$\cdot$s, compared to the BODE's lower value of 0.6~Pa$\cdot$s, as illustrated in Figure~\ref{fig:uncertainty_2500}(b). The epistemic uncertainty is also related to data scarcity or distribution, which may explain why the epistemic uncertainty cannot be reduced to zero. Aleatoric uncertainty is significantly overestimated by the baseline ensemble, reaching a value of 2.5~Pa$\cdot$s, while the BODE predicts near-zero aleatoric uncertainty, as shown in Figure~\ref{fig:uncertainty_2500}(c). Since the CFD data are noise-free, the BODE's near-zero aleatoric uncertainty aligns well with expectations.

\begin{figure}[H]
    \centering
    \includegraphics[width=\textwidth]{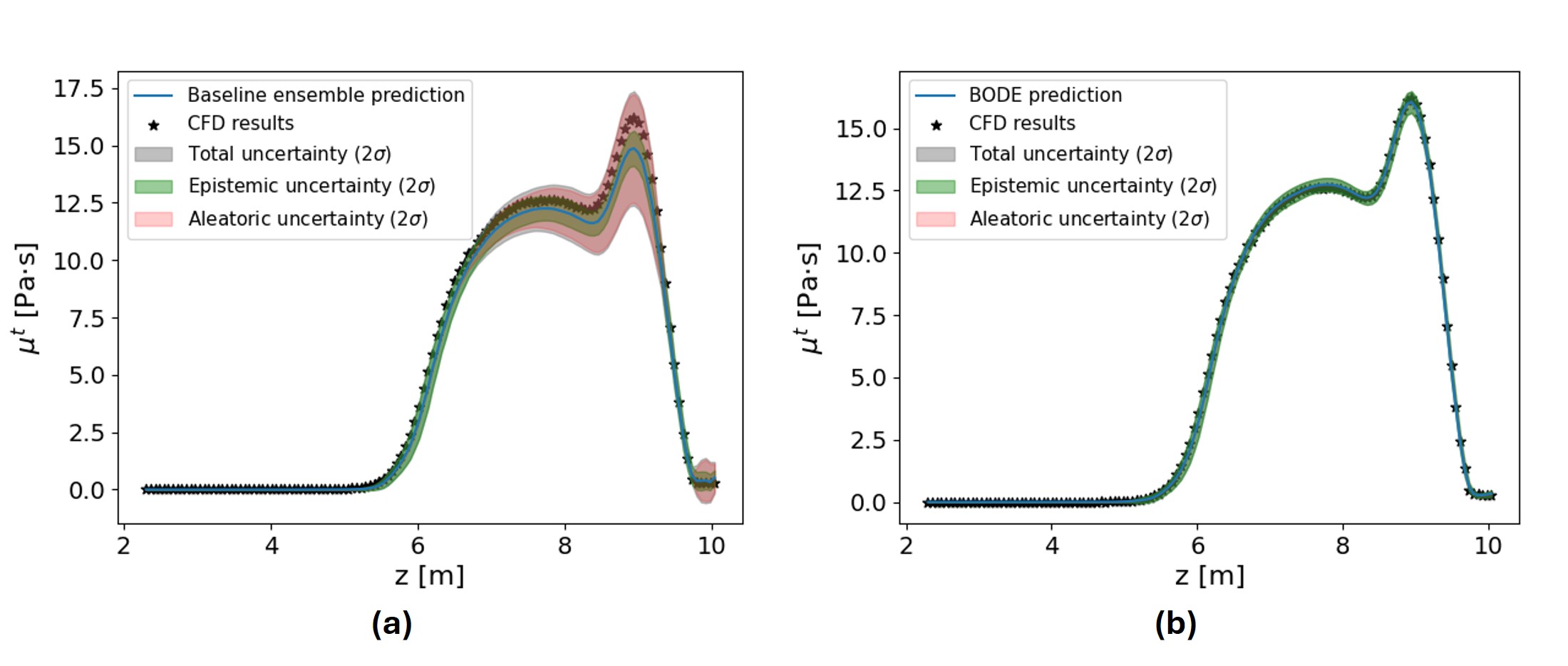}
    \caption{Prediction and uncertainties along the symmetry line, where error bars represent the added noise for (a) baseline ensemble; (c) BODE.}
    \label{fig:pred_line}
\end{figure}

A quantitative comparison along the symmetry line further supports these findings. As shown in Figure~\ref{fig:pred_line}, the predictions and estimated uncertainties from both the baseline ensemble and BODE models are compared against the CFD results. The baseline ensemble model exhibits broader total uncertainty bounds at higher $z$ values near the top of the tank, primarily due to increased aleatoric uncertainty in this region (Figure~\ref{fig:pred_line}.a). In contrast, the BODE model shows narrower uncertainty bounds, as depicted in Figure~\ref{fig:pred_line}.b, indicating that the BODE method provides more precise predictions and better captures data noise. In summary, both epistemic and aleatoric uncertainties are more pronounced in the baseline ensemble model compared to the BODE model.
\subsection{Impact of Data Noise}

\begin{figure}[H]  
    \centering
    \includegraphics[width=\linewidth]{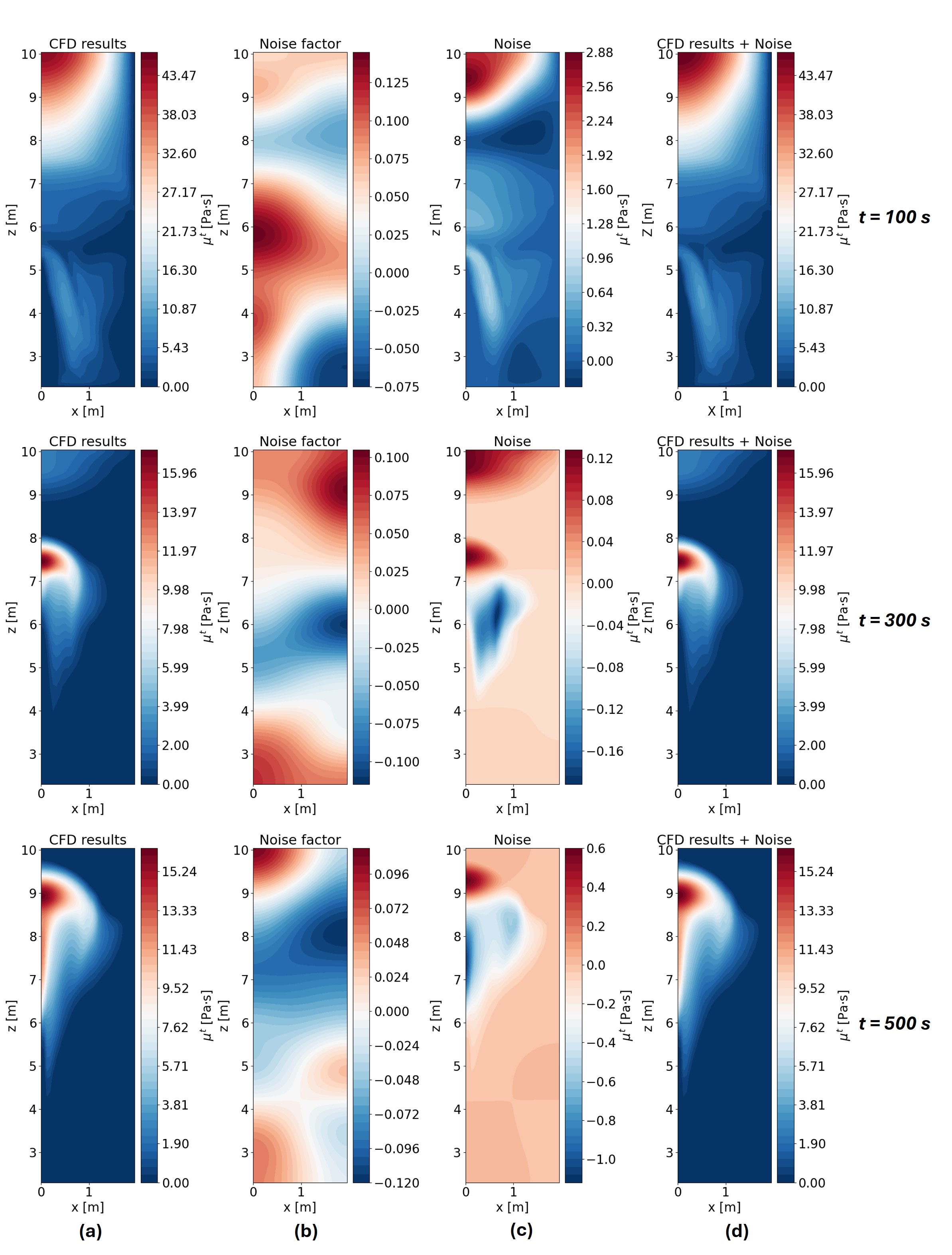}  
    \caption{CFD results for a noise factor with 5\% standard deviation, showing the original CFD results, noise factor, generated noise, and the final CFD results with noise applied for (a) time step $t = 100$ $s$, (b) $t = 300$ $s$, and (c) time step $t = 500$ $s$.}
    \label{fig:noise_5_per}
\end{figure}

\begin{figure}[H]  
    \centering
    \includegraphics[width=\linewidth]{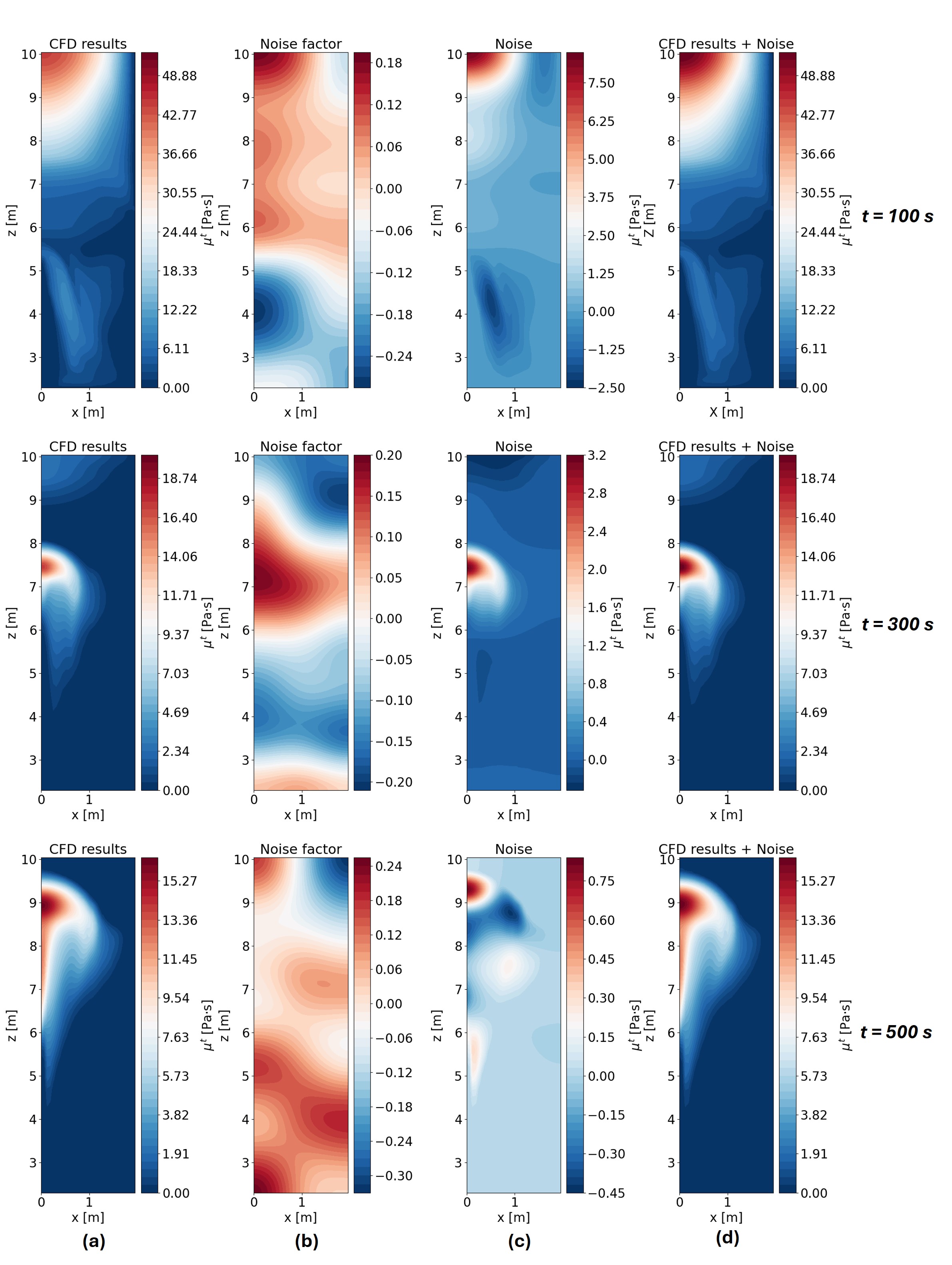}  
    \caption{CFD results for a noise factor with 10\% standard deviation, showing the original CFD results, noise factor, generated noise, and the final CFD results with noise applied for (a) time step $t = 100$ $s$, (b) $t = 300$ $s$, and (c) time step $t = 500$ $s$.}
    \label{fig:noise_10_per}
\end{figure}

In the previous section, we demonstrated that the BODE successfully predicts near-zero aleatoric uncertainty in a noise-free environment. To evaluate the BODE's ability to quantify uncertainties and make predictions in noisy conditions, we introduce Gaussian noise to the eddy viscosity fields derived from the CFD results. This approach simulates realistic variability often present in engineering scenarios. Figures~\ref{fig:noise_5_per} and \ref{fig:noise_10_per} display the original CFD results, the noise factor, the generated noise, and the resulting CFD fields with applied noise, corresponding to noise factors with 5\% and 10\% standard deviations for time steps $t = 100$~s, $t = 300$~s, and $t = 500$~s, respectively. The addition of Gaussian noise allows us to assess the BODE's accuracy in UQ under noisy conditions.

The original CFD results depict the eddy viscosity $\mu^t$ (in Pa$\cdot$s) from the simulations. These figures are identical for the same time step in Figures~\ref{fig:noise_5_per}(a) and \ref{fig:noise_10_per}(a), but they are plotted on different scales to facilitate comparison with the perturbed noise. At earlier time steps, the eddy viscosity $\mu^t$ is notably higher, with significant changes in its pattern across different time steps. For instance, at $t = 100$~s, there is a concentration of high eddy viscosity near the top of the domain, whereas at $t = 300$~s, this region shifts lower compared to $t = 100$~s.

A dimensionless noise factor, shown in Figures~\ref{fig:noise_5_per}(b) and \ref{fig:noise_10_per}(b), is applied to these results, introducing variability through added or subtracted noise. This noise factor $\epsilon \sim \mathcal{N}(0, \sigma^2)$ has $\sigma$ values corresponding to 5\% and 10\% standard deviations. A Gaussian filter is applied to the noise factor, defined by Equation~\ref{eq:filter}, to smooth it. The noise factor ranges from $[-3\sigma, 3\sigma]$, translating to $[-0.15, 0.15]$ for a 5\% standard deviation and $[-0.3, 0.3]$ for a 10\% standard deviation. The generated noise, being proportional to $\mu^t$, can be positive or negative, with a 10\% standard deviation introducing more pronounced noise than a 5\% standard deviation.

Since the noise is calculated by multiplying the noise factor with the CFD results, as in Equation~\ref{eq:final_noise}, it aligns with the spatial distribution of the eddy viscosity field, resulting in more pronounced noise in areas with higher eddy viscosity, as demonstrated in Figures~\ref{fig:noise_5_per}(c) and \ref{fig:noise_10_per}(c). The noise perturbations follow the spatial patterns of the eddy viscosity field, where regions of higher $\mu^t$ exhibit greater noise.

\begin{figure}[H]  
    \centering
    \includegraphics[width=1\linewidth]{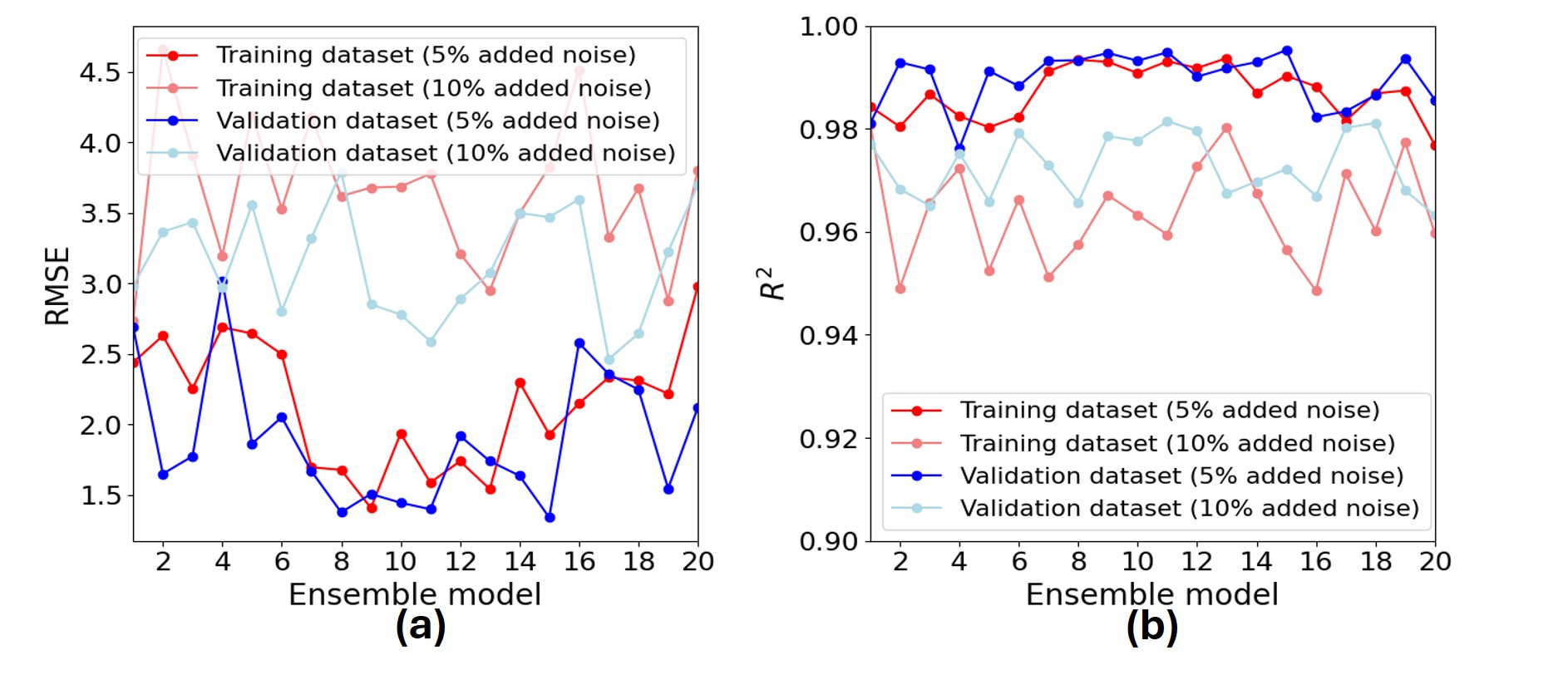}
    \caption{Comparison between baseline ensemble and BODE in terms of (a) RMSE and (b) $R^2$ for training and validation datasets with added noise (5\% and 10\% standard deviations).}
    \label{fig:RMSE_R2_noise}
\end{figure}

We apply the BODE to the CFD results with added Gaussian noise at standard deviations of 5\% and 10\%. The RMSE and $R^2$ metrics for these cases are shown in Figure~\ref{fig:RMSE_R2_noise}. The added noise affects prediction accuracy: with 5\% standard deviation noise, the average RMSE for the training and validation datasets are 2.15~Pa$\cdot$s and 1.90~Pa$\cdot$s, respectively, while the average $R^2$ values are 0.987 and 0.990. For the 10\% noise case, the BODE model shows higher RMSE and lower $R^2$: the average RMSE for the training and validation datasets are 3.64~Pa$\cdot$s and 3.15~Pa$\cdot$s, respectively, with corresponding average $R^2$ values of 0.964 and 0.973.

Despite differences in performance metrics between the 5\% and 10\% noise cases, the BODE effectively demonstrates its capability to predict eddy viscosity fields even with added Gaussian noise.

\begin{figure}[H]  
    \centering
    \includegraphics[width=\linewidth]{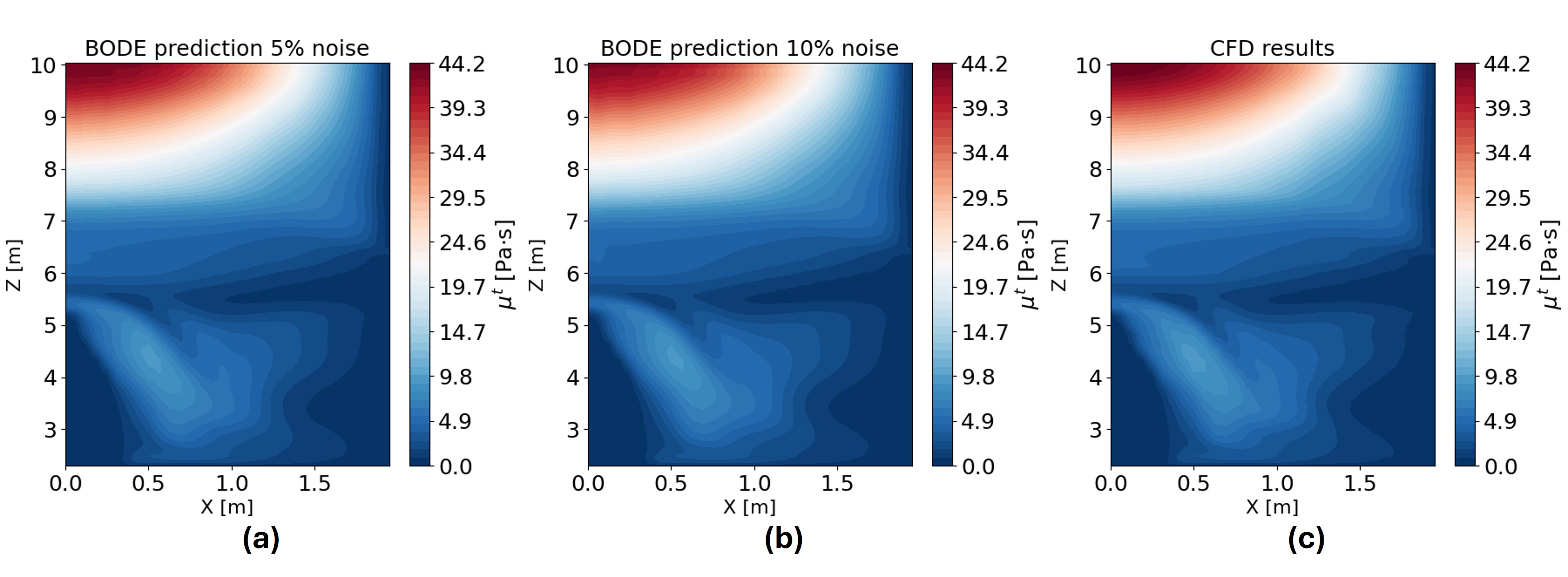}
    \caption{Comparison of eddy viscosity $\mu^t$ at time step $t = 100$~s: (a) BODE with 5\% standard deviation noise; (b) BODE with 10\% standard deviation noise; (c) Original CFD results.}
    \label{fig:prediction_500_noise}
\end{figure}

\begin{figure}[H]  
    \centering
    \includegraphics[width=\linewidth]{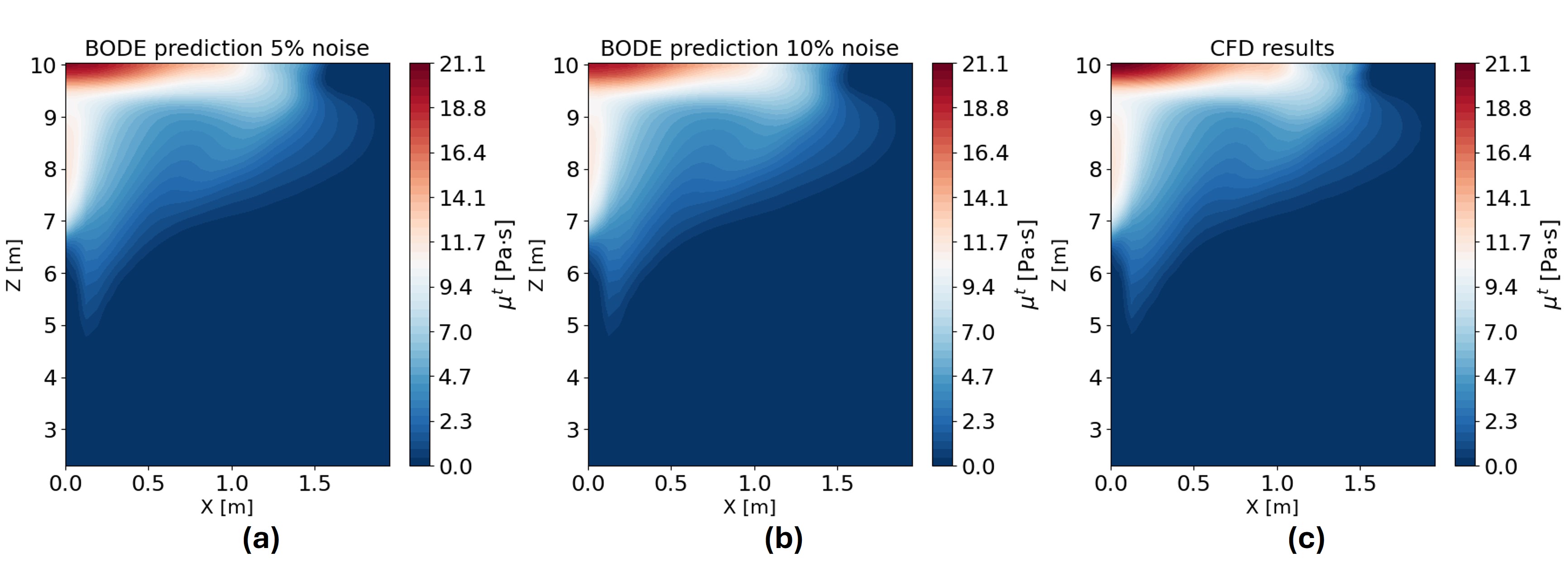}
    \caption{Comparison of eddy viscosity $\mu^t$ at time step $t = 600$~s: (a) BODE with 5\% standard deviation noise; (b) BODE with 10\% standard deviation noise; (c) Original CFD results.}
    \label{fig:prediction_3000_noise}
\end{figure}

BODE predictions at time steps $t = 100$~s and $t = 600$~s with noise factors of 5\% and 10\% are illustrated in Figures~\ref{fig:prediction_500_noise} and \ref{fig:prediction_3000_noise}. Figures~\ref{fig:prediction_500_noise}(a) and \ref{fig:prediction_3000_noise}(a) depict BODE mean predictions with a noise factor of 5\% standard deviation, while Figures~\ref{fig:prediction_500_noise}(b) and \ref{fig:prediction_3000_noise}(b) show BODE mean predictions with a 10\% standard deviation noise factor. For comparison, Figures~\ref{fig:prediction_500_noise}(c) and \ref{fig:prediction_3000_noise}(c) present the original CFD results.

The BODE model successfully captures both the global features and local variations of the eddy viscosity fields with high accuracy, even in the presence of noise. As described by Equation~\ref{eq:loss_nll}, the negative log-likelihood function predicts both the mean and noise (standard deviation). Through optimization, the neural network effectively filters out the noise, accurately capturing the underlying mean of the eddy viscosity field. This outcome demonstrates the robustness of the BODE model in handling noisy data while maintaining high prediction accuracy.

\begin{figure}[H]  
    \centering
    \includegraphics[width=0.74\linewidth]{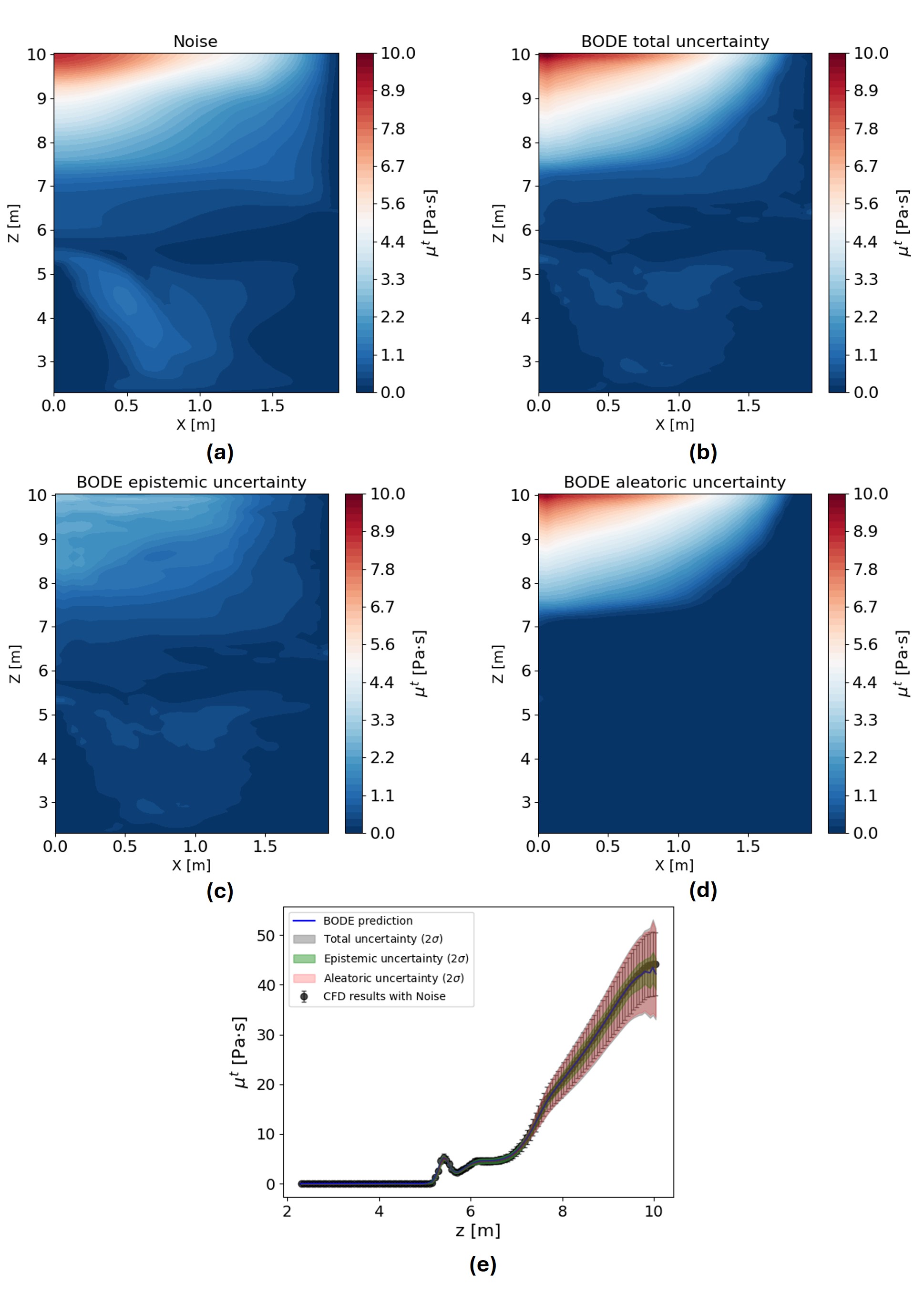}
    \caption{Uncertainty analysis at time step $t = 100$~s with 5\% standard deviation noise: (a) Total uncertainty; (b) Added noise; (c) Epistemic uncertainty; (d) Aleatoric uncertainty; (e) Uncertainties along the symmetry line, where error bars represent the added noise.}
    \label{fig:uncertainty_500_5per}
\end{figure}

\begin{figure}[H]  
    \centering
    \includegraphics[width=0.74\linewidth]{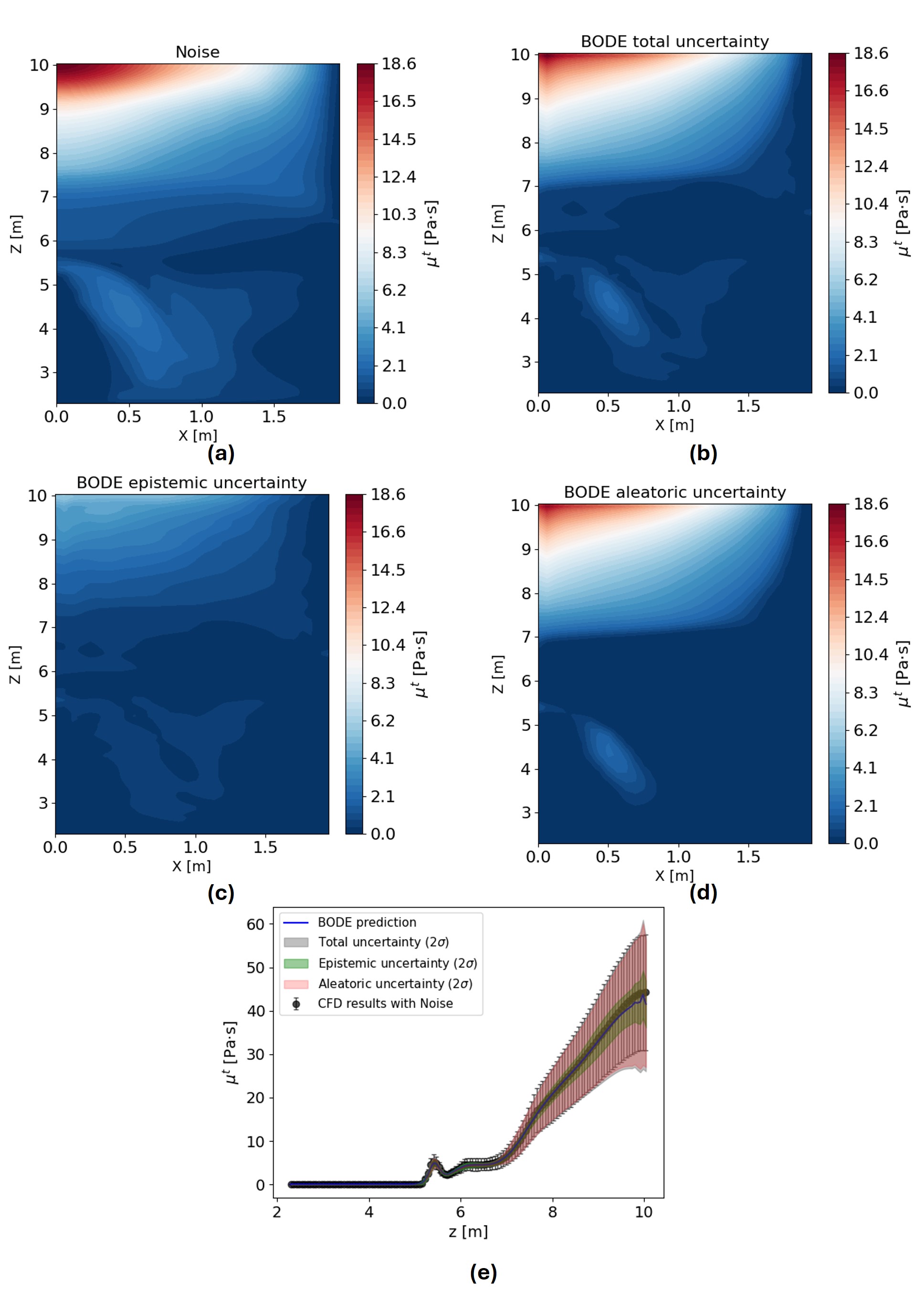}
    \caption{Uncertainty analysis at time step $t = 100$~s with 10\% standard deviation noise: (a) Total uncertainty; (b) Added noise; (c) Epistemic uncertainty; (d) Aleatoric uncertainty; (e) Uncertainties along the symmetry line, where error bars represent the added noise.}
    \label{fig:uncertainty_500_10per}
\end{figure}

The UQ analysis with added 5\% and 10\% standard deviation noise is shown in Figures~\ref{fig:uncertainty_500_5per} and \ref{fig:uncertainty_500_10per}, corresponding to the time step $t = 100$~s. These figures compare the total, aleatoric, and epistemic uncertainties estimated by the BODE with the original added noise. The added noise is illustrated in Figures~\ref{fig:uncertainty_500_5per}(b) and \ref{fig:uncertainty_500_10per}(b). The BODE effectively captures the global features of the noise, as shown in the total uncertainty plots (Figures~\ref{fig:uncertainty_500_5per}(a) and \ref{fig:uncertainty_500_10per}(a)); however, some discrepancies are observed in the local values.

The aleatoric uncertainty, which represents the inherent data noise, is captured by the BODE and is concentrated in regions with higher eddy viscosity values, as demonstrated in Figures~\ref{fig:uncertainty_500_5per}(d) and \ref{fig:uncertainty_500_10per}(d). The epistemic uncertainty, reflecting model uncertainty, is also present and spreads across regions of non-zero eddy viscosity $\mu^t$, as illustrated in Figures~\ref{fig:uncertainty_500_5per}(c) and \ref{fig:uncertainty_500_10per}(c). Ideally, the noise should be captured as aleatoric uncertainty, while the epistemic uncertainty should be minimized. While the BODE captures substantial noise as aleatoric uncertainty, it struggles to represent finer spatial gradients of the noise, which are partially captured as epistemic uncertainty. Increasing the number of ensemble models could lead to more accurate estimations of the noise in these regions.

The BODE's estimation of the uncertainty profile along the symmetry line shows good agreement with the original noise data, as evidenced in Figures~\ref{fig:uncertainty_500_5per}(e) and \ref{fig:uncertainty_500_10per}(e). The BODE's ability to estimate noise is spatially variable across the domain and performs better in regions with higher eddy viscosity $\mu^t$ values for both noise levels. The BODE's capability to predict uncertainty is not significantly impacted by the increase in noise levels, as both cases exhibit similar accuracy and uncertainty patterns. In the case of a 10\% noise factor, the BODE estimates nonzero aleatoric uncertainty localized in the lower regions, which is less pronounced in the 5\% case. This observation indicates that the BODE is more sensitive to higher noise levels when capturing noise as aleatoric uncertainty.

\subsection{Discussion}
The performance of a DNN is significantly influenced by its hyperparameters. Additionally, overfitting can limit the NN's ability to generalize to new data. By using BO with a validation dataset rather than a training dataset, the risk of overfitting is minimized. The objective function, in this case the RMSE, is represented as a function of the hyperparameters in the surrogate model. Selecting ensemble members from one or more surrogate models results in enhanced accuracy and UQ.

The BODE results in better alignment of aleatoric uncertainty with actual data noise and a reduction of epistemic uncertainty, indicating increased model confidence. Moreover, the BODE method is flexible and applicable to other data-driven models and NNs.

This study emphasizes the importance of optimizing the NN for both prediction accuracy and UQ. A robust optimization method is necessary, and BO is one of the most effective techniques. Current developments focus heavily on the implementation of NNs in various engineering and medical fields, but optimization plays a crucial role in achieving reliable predictions and UQ.

One limitation of this work is that the DE approach assumes the data follow a Gaussian distribution. While many engineering applications and measurements are well-described by a Gaussian distribution, this assumption could be a limitation in some cases. In this study, a Gaussian noise factor is generated and then smoothed using a Gaussian filter. The result of Gaussian convolution is Gaussian. However, a max function is applied to eliminate any negative values of the eddy viscosity $\mu^t$, which may distort the Gaussian distribution. Analyzing a case where the output parameter can take negative values could provide further insights.

Another aspect of the current research is the use of RMSE instead of the NLL for optimization. The decision to use RMSE is to avoid conflicting terms in the NLL loss function, where the variance appears in both the denominator of the first term and the numerator of the second term, as shown in Equation~\ref{eq:loss_nll}. Studying the optimization using the NLL and its effect on the UQ could be beneficial.

Additionally, the BODE captures finer spatial gradient noise in the form of epistemic uncertainty. Therefore, the total uncertainty provides a good representation of the noise if a sufficient number of optimized ensemble members are used. In this study, twenty ensemble members are used. Investigating the effect of increasing the number of ensemble members could enhance the results. However, capturing the noise in regions with smaller gradient variations as epistemic uncertainty raises questions about the separation of the total uncertainty into aleatoric and epistemic components.

Future research could include assessing the BODE on different NN architectures. While this research focuses on discriminative AI, it could also be expanded to generative AI. Additionally, increasing the number of iterations in the Sobol sequence and analyzing its effect on the optimization could provide further improvements. The impact of data distribution on epistemic uncertainty, which is related to data scarcity or distribution, should also be investigated.

\section{Conclusions}
\label{sec:conclusions}

In this work, we proposed a novel method that combines Bayesian Optimization (BO) with Deep Ensembles (DE), referred to as BODE, to enhance both prediction accuracy and uncertainty quantification (UQ) in Deep Neural Networks (DNNs). Accurate predictions with reliable UQ are essential in risk-sensitive fields such as nuclear safety, where decision-making relies heavily on the confidence in model predictions. Our approach addresses the limitations of traditional DE methods, which often exhibit overestimated aleatoric uncertainty and discrepancies between model predictions and actual data, especially when constructed by retraining the same neural network multiple times with randomly sampled initializations.

By integrating BO into the ensemble construction process, BODE advances the state-of-the-art in UQ for DNNs. Unlike conventional DEs, which rely on varying initializations of the same architecture, BODE leverages BO to optimize hyperparameters for each ensemble member, resulting in a diverse set of models with different architectures and training dynamics. This diversity enhances the ensemble's ability to capture epistemic uncertainty and reduces the misinterpretation of model uncertainty as aleatoric uncertainty.

We applied the BODE method to a case study involving eddy viscosity prediction in sodium fast reactor thermal stratification modeling. By optimizing the hyperparameters of a Densely connected Convolutional Neural Network (DCNN) through BO, we achieved significant improvements over a manually tuned baseline ensemble (BE). In a noise-free environment, BODE reduced the total uncertainty by approximately nine times compared to the BE, primarily due to a substantial reduction in overestimated aleatoric uncertainty. Specifically, BODE estimated aleatoric uncertainty close to zero, accurately reflecting the absence of noise, and demonstrated higher predictive accuracy with an average RMSE of 0.77~Pa$\cdot$s and an $R^2$ value of 0.997.

Furthermore, when Gaussian noise with standard deviations of 5\% and 10\% was introduced into the eddy viscosity data, BODE effectively captured the data noise and provided uncertainty estimates that aligned with the actual noise levels. The method accurately fitted the noisy data without overfitting, demonstrating robustness in the presence of data noise. Although some fine spatial variations in noise were partially captured as epistemic uncertainty, the total uncertainty provided a reliable representation of the data noise.

These results demonstrate that combining BO with DE is an effective strategy for improving both the predictive performance and uncertainty estimation of DNN models. By optimizing hyperparameters and leveraging the diversity of ensemble members, BODE enhances the model's ability to generalize and accurately quantify uncertainties, making it a valuable tool for applications requiring high reliability and confidence in predictions.

In the context of nuclear safety, where accurate modeling of complex phenomena like thermal stratification is critical, the BODE method offers a promising approach to enhance predictive capabilities while providing reliable UQ. This can contribute to better risk assessment, improved safety margins, and informed decision-making. While the study focused on a specific application, the methodology is flexible and can be extended to other domains and neural network architectures.

Future work may involve applying BODE to experimental data, where inherent noise is present, allowing for direct comparison between BODE's uncertainty estimates and experimental uncertainty analyses. Additionally, exploring the impact of ensemble size on epistemic uncertainty and investigating alternative loss functions or optimization strategies could provide further insights into improving UQ in DNN models.

In conclusion, the BODE method advances the state of the art in combining optimization techniques with ensemble learning for UQ. It addresses key challenges in data-driven modeling by reducing overestimated uncertainties and enhancing prediction accuracy, thereby contributing to the development of more reliable and robust models in safety-critical applications.

\section*{Data Availability}
The code and data related to this work will be made publicly available on GitHub after the official publication of this manuscript.

\section*{Acknowledgements}

This work is supported by the US Department of Energy Office of Nuclear Energy Distinguished Early Career Program under contract number DE-NE0009468.

\bibliographystyle{elsarticle-num}   

\bibliography{refs} 
\end{document}